\documentclass{article}

\usepackage[english]{babel}

\usepackage[letterpaper,top=2cm,bottom=2cm,left=3cm,right=3cm,marginparwidth=1.75cm]{geometry}

\usepackage{amsmath}
\usepackage{graphicx}
\usepackage[colorlinks=true, allcolors=blue]{hyperref}

\usepackage[utf8]{inputenc} 
\usepackage[T1]{fontenc}    
\usepackage{hyperref}       
\hypersetup{colorlinks=true}
\usepackage{url}            
\usepackage{booktabs}       
\usepackage{amsfonts}       
\usepackage{nicefrac}       
\usepackage{microtype}      
\usepackage{xcolor}         
\usepackage[ruled,vlined,linesnumbered,commentsnumbered]{algorithm2e}
\usepackage{algorithmicx}  
\usepackage{algpseudocode}  

\usepackage{microtype}
\usepackage{multirow}
\usepackage{subfigure}
\usepackage{booktabs} 
\usepackage{amsmath}
\usepackage{mathtools}
\usepackage{enumitem}
\usepackage[capitalize,noabbrev]{cleveref}
\usepackage{verbatim}

\usepackage{CJKutf8}
\usepackage{enumitem}
\usepackage{textcomp}
\usepackage{duckuments}
\usepackage{bbding}
\usepackage{pifont}
\usepackage{color}

\definecolor{MyDarkBlue}{rgb}{0,0.5,1}
\definecolor{MyDarkGreen}{rgb}{0.02,0.6,0.02}
\definecolor{MyDarkRed}{rgb}{0.8,0.02,0.02}
\definecolor{MyDarkOrange}{rgb}{0.40,0.2,0.02}
\definecolor{MyPurple}{RGB}{111,0,255}
\definecolor{MyRed}{rgb}{1.0,0.0,0.0}
\definecolor{MyGold}{rgb}{0.75,0.6,0.12}
\definecolor{MyDarkgray}{rgb}{0.66, 0.66, 0.66}

\newcommand{\modelname}{MetaTrader}
\newcommand{\cmark}{\ding{51}}
\newcommand{\xmark}{\ding{55}}

\title{Your Offline Policy is Not Trustworthy: Bilevel Reinforcement Learning for Sequential Portfolio Optimization}
\author{Haochen Yuan$^{1\dagger}$,\quad
	Minting Pan$^{1\dagger}$,\quad
    Yunbo Wang$^{1\ast}$,\quad
    Siyu Gao$^1$,\\
    Philip S. Yu$^2$,\quad
    Xiaokang Yang$^1$}

\date{\small $^1$ MoE Key Lab of Artificial Intelligence, AI Institute, Shanghai Jiao Tong University, China
\\
\small $^2$ Department of Computer Science, University of Illinois Chicago, USA
\\
$^\dagger$These authors contributed equally to this work.
\\
$^\ast$Corresponding author. Email: yunbow@sjtu.edu.cn.
}

\begin{document}
\maketitle

\begin{abstract}
Reinforcement learning (RL) has shown significant promise for sequential portfolio optimization tasks, such as stock trading, where the objective is to maximize cumulative returns while minimizing risks using historical data. However, traditional RL approaches often produce policies that merely ``\textit{memorize}'' the optimal yet impractical buying and selling behaviors within the fixed dataset. These offline policies are less generalizable as they fail to account for the non-stationary nature of the market. Our approach, MetaTrader, frames portfolio optimization as a new type of partial-offline RL problem and makes two technical contributions. First, MetaTrader employs a bilevel learning framework that explicitly trains the RL agent to improve both in-domain profits on the original dataset and out-of-domain performance across diverse transformations of the raw financial data. Second, our approach incorporates a new temporal difference (TD) method that approximates \textit{worst-case} TD estimates from a batch of transformed TD targets, addressing the value overestimation issue that is particularly challenging in scenarios with limited offline data. Our empirical results on two public stock datasets show that MetaTrader outperforms existing methods, including both RL-based approaches and traditional stock prediction models.

\end{abstract}

\section{Introduction}



\noindent
Portfolio optimization refers to the process of selecting the best mix of assets (such as stocks, bonds, or other investment vehicles) to achieve a specific financial goal, often maximizing return while minimizing risk. The objective is to create a portfolio that offers the most efficient balance between risk and return, based on the investor’s preferences, constraints, and market conditions.
In traditional methods, portfolio optimization often uses historical data (such as past returns and covariances between assets) to determine the optimal asset allocation. However, modern reinforcement learning (RL) approaches aim to optimize portfolios dynamically, accounting for changing market conditions over time~\cite{deng2016deep,ye2020reinforcement,briola2021deep,liu2021finrl,kumar2023deep,gao2023stockformer}.

Recent advances in RL-based trading, such as StockFormer~\cite{gao2023stockformer}, have demonstrated superior performance compared to simpler strategies that combine stock prediction models~\cite{li2018stock,xu2018stock,wang2021hierarchical,zheng2023deep} with fixed trading policies---such as buying stocks with the highest predicted future gains and holding them for a predefined period. 
These RL approaches commonly employ advanced deep learning models to extract meaningful features from the noisy market data, \textit{e.g.}, stock prices, trading volumes, and financial news. These extracted features are then used as inputs for RL algorithms, which are typically designed to maximize the expected total payoff within the constraints of the offline training data.

\begin{figure}[t]
\begin{center}
\centerline{\includegraphics[width=\columnwidth]{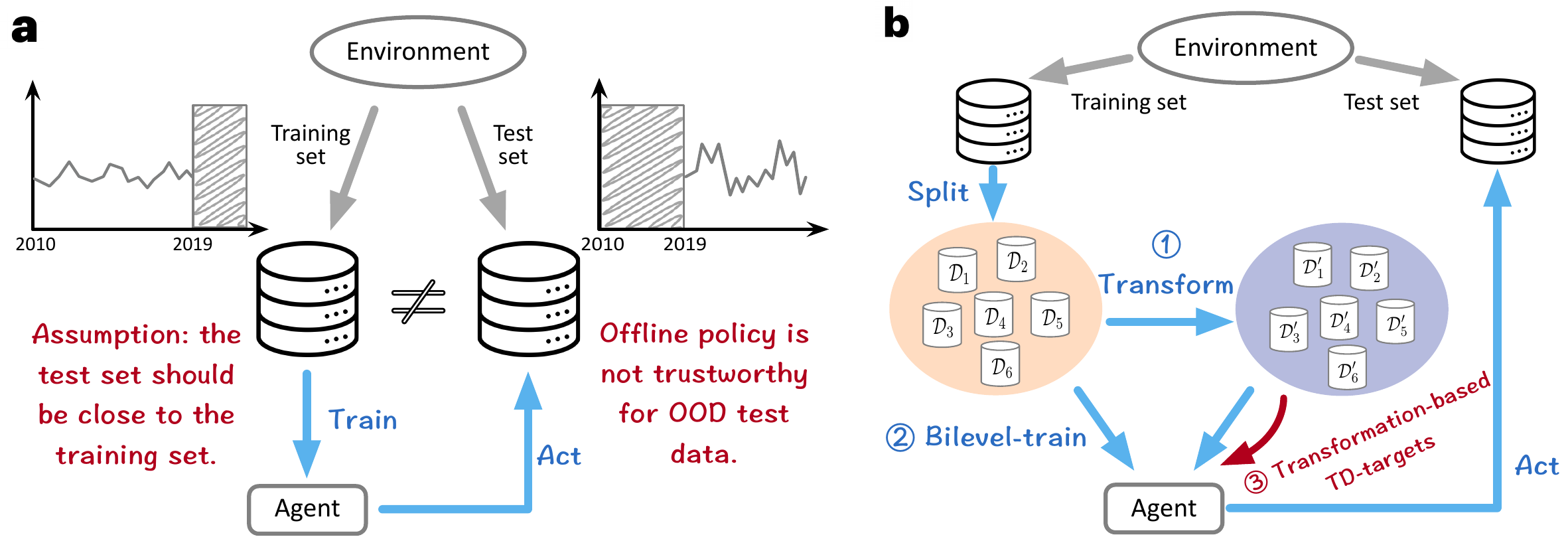}}
\vspace{-5pt}
\caption{\textbf{A comparison of \modelname{} and existing RL-based trading methods.} \textbf{a,} Existing RL-for-finance methods typically adopt an offline training setup rather than online RL, causing them to struggle with the generalization-optimality dilemma, a common challenge in the inherently non-stationary financial market. \textbf{b,} \modelname{} tackles this paradox through: (1) specialized data transformations to simulate OOD financial data, (2) a bilevel RL framework that explicitly optimizes both in- and out-of-domain performance across diverse transformations, and (3) a novel TD learning method that conservatively estimates state-action values by approximating the minimum TD targets generated from a batch of data transformations.
}
\label{fig:intro}
\end{center}
\vspace{-20pt}
\end{figure}

\vspace{-5pt}
\paragraph{Generalization-optimality dilemma.} 
However, most existing RL-for-finance methods apply standard RL algorithms to find an optimal policy from a previously collected static dataset, which bears algorithmic difficulties due to function approximation errors from out-of-distribution (OOD) data points.
Since the RL agent cannot actively explore the rapidly evolving financial market, it tends to overfit the historical data and simply memorize the ``optimal'' offline policy---transactions that yield the highest profits within the dataset---even though such a policy may not be generalizable outside the dataset's scope. 
This raises a crucial yet under-explored question: \textit{How can we learn more robust trading policies that can jointly handle the in-domain optimality\footnote{In-domain optimality refers to achieving the best possible financial outcomes, measured by maximum rewards, within the given historical data. It emphasizes the agent's performance within the specific dataset it was trained on.} and out-of-domain generalizability?}

The generalization-optimality dilemma can also be a fitting description for the exploration-exploitation dilemma in our offline RL formulation of the portfolio optimization problem.
Generalization refers to an agent's ability to apply knowledge gained from a fixed dataset to unseen market conditions. This capability is crucial for tasks where the training dataset is incomplete or significantly diverges from the test data distribution.
However, excessive generalization risks the agent straying into OOD regions where the model's estimations of state-action values become unreliable, potentially resulting in poor performance or unsafe behaviors.
Optimality, on the other hand, focuses on maximizing rewards by closely following the optimal actions within the training dataset, ensuring reliable performance in familiar scenarios.
Yet, an overemphasis on in-domain optimality can hinder the agent's ability to explore potentially superior policies beyond the dataset's boundaries. This limitation may result in suboptimal behavior in highly non-stationary, evolving environments where broader exploration could reveal more effective solutions.

\vspace{-5pt}
\paragraph{Overview of \modelname{}.}
In this paper, we introduce \modelname{}, an early study on bilevel optimization of actor-critic methods in stock trading, formulated as a \textit{partial-offline} RL problem with decoupled state branches.
The core idea of \modelname{} goes beyond maximizing expected total rewards on current trajectories, learning policies that also perform effectively on OOD financial data. 
As shown in Figure \ref{fig:intro}b, we enhance existing RL-based trading methods in two key areas: a bilevel RL framework and a novel temporal difference (TD) learning approach.

A primary contribution of our work is to enhance the generalization capability of the policy from both the data transformation and algorithmic perspectives, which are closely interconnected.
From a data perspective, we introduce specific data transformation methods designed to simulate OOD samples. These transformations focus on different factorized components of the time series data, including \textit{short-term} randomness, \textit{long-term} trends, and \textit{multi-scale} correlations.

From an algorithmic perspective, we propose a novel actor-critic method based on bilevel optimization.
Incorporating bilevel gradient updates helps prevent the agent from overfitting to the historical distribution by explicitly evaluating the hypothetical model parameters on the transformed OOD market data. This training strategy ensures the model does not simply memorize the optimal policy based on specific patterns in the training data.


Another contribution of our work is a novel TD learning method that conservatively estimates state-action values by approximating the minimum TD targets generated from a batch of data transformations.
This approach seeks to enhance the generalizability of policies learned from offline data to mitigate the value overestimation issue, which is particularly severe when there are significant discrepancies between the training and test distributions of non-stationary market data.
We empirically demonstrate that the pronounced distributional shift makes the existing conservative offline RL methods, such as CQL~\cite{kumar2020conservative} and IQL~\cite{kostrikov2021offline}, inadequate for tackling RL-for-finance tasks effectively.
Specifically, we modify TD learning by constructing an ensemble of TD targets, separately computing the next-step Q-values for both the original data and its transformations. We then use the minimum Q-value among them as the TD target to train the current-step value estimate. 
%
Unlike previous ensemble-based Q-learning methods, which use the multiple target Q-networks to compute ensemble value regularization, our method relies on a single target Q-network and derives the \textit{worst-case} Q-value through a diverse set of transformed data.

Our approach significantly outperforms existing RL-for-finance methods on the CSI-300 and the NASDAQ-100 stock datasets, achieving superior cumulative returns and Sharpe ratios. This highlights its ability to effectively balance trading profits and risks.
By addressing critical challenges such as policy overfitting and value overestimation, our method offers a more reliable and adaptive solution for financial trading in real-world, non-stationary environments.
Additionally, the techniques we propose provide a generalizable framework that can be extended to a wide range of non-stationary decision-making scenarios, such as autonomous driving and power systems management, where adaptability to dynamic environments is crucial for success.

\section{Problem Formulation: Partial-Offline Reinforcement Learning}
\label{sec:prob}


\begin{figure}[t]
\begin{center}
\centerline{\includegraphics[width=\columnwidth]{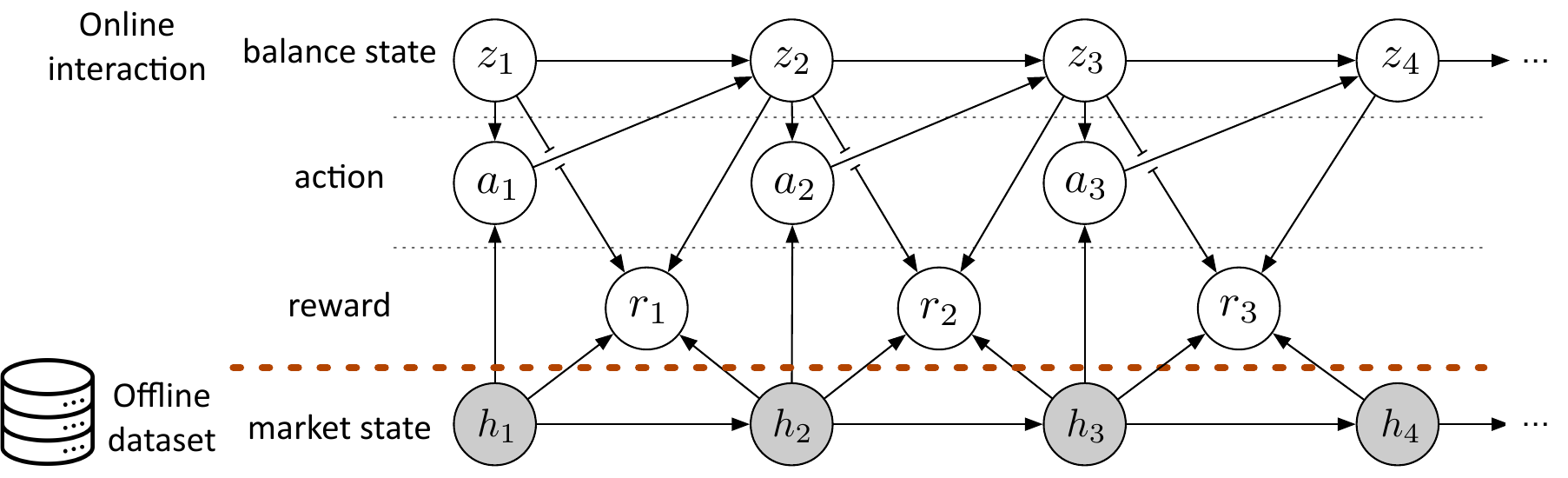}}
\vspace{-5pt}
\caption{\textbf{The MDP in the partial-offline RL setup for sequential portfolio optimization.}
    The MDP consists of decoupled pairs of action-free \textit{market states} and action-dependent \textit{balance states}, with market states restricted to the offline training set.
    Unlike standard offline RL, where no new rewards are accessible during policy optimization, the partial-offline setup allows the agent to interact with the fixed training set, exploring different policies and collecting new reward feedback.
}
\label{fig:setting}
\end{center}
\vspace{-20pt}
\end{figure}

We introduce a novel formulation of sequential portfolio optimization as a ``\textit{partial-offline}'' RL problem, with the key distinction from the standard offline RL setup explained later.

As shown in Figure \ref{fig:setting}, the Markov decision process (MDP) in the context of stock trading can be described as an $8$-tuple $(\mathcal O, \mathcal A, \mathcal H, \mathcal Z, P_h, P_z, R, \gamma)$:

\vspace{-5pt}
\paragraph{Observation space ($\mathcal O$).} The raw data includes: (1) $o_t^\text{price} \in \mathbb{R}^{T \times |S|\times 5}$: Daily \textit{open, close, high, low} stock prices, and trading \textit{volumes} for the previous $T$ days. $|S|$ is the total number of stocks. (2) $o_t^\text{stat} \in \mathbb{R}^{|S|\times I}$: $I$ technical indicators that reflect the temporal trends of stock prices. (3) A covariance matrix $o_t^\text{cov}$ that measures the correlations between historical daily closing prices of all stocks.
In our partial-offline RL setup, only a finite set of observation data is accessible.

\vspace{-5pt}
\paragraph{Action space ($\mathcal A$).}
We use a continuous action space $a_t \in \mathbb{R}^{|S|}$, where each component represents the number of shares to buy, hold, or sell for each asset. To simulate real-world trading, we discretize $a_t$ into several intervals, such as $100,200,\ldots$ shares when deploying the agent for testing.

\vspace{-5pt}
\paragraph{Decoupled state space ($\mathcal H$, $\mathcal Z$) and state transitions ($P_h$, $P_z$).}

We decouple the state space into two components: $\mathcal S = (\mathcal H, \mathcal Z)$. Here, $\mathcal H$ is the \textit{market state space} represented by the embeddings from the observed financial data, while $\mathcal Z$ is the \textit{balance state space} that models the balance sheet.
The \textit{market state} $h_t$ is composed of three types of latent states $(h_t^\text{relat}$, $h_t^\text{long}$, $h_t^\text{short})$ generated from $o_t^\text{price}$, $o_t^\text{stat}$ and $o_t^\text{cov}$ using encoding networks. Please refer to Eq. \eqref{eq:stockformer} for details.
The \textit{balance state} $z_t \in \mathbb{R}^{|S|+1}$ represents the total account balance and holding amount of each trading asset.

Since individual buying and selling actions typically have minimal impact on market dynamics, market state transitions are largely \textit{action-free} whereas balance state transitions are \textit{action-dependent}. Therefore, we define the state transition probabilities as $P_h(h_{t+1}|h_t)$ for market states and $P_z(z_{t+1}|z_t, a_t)$ for balance states.
In partial-offline RL, market states are limited to the offline training set, while the agent can explore various actions that lead to new balance states.

\vspace{-5pt}
\paragraph{Reward function ($R$) and discount factor ($\gamma$).} The immediate reward is defined as the daily portfolio return ratios: $r_t={R}(h_{t:t+1}, z_{t:t+1})$, where $z_{t+1}$ is dependent on $a_t$.
$\gamma$ is the reward discount factor that determines how much the RL agents care about rewards in the distant future.

\vspace{-5pt}
\paragraph{Distinctions from standard offline RL setups.}
In standard offline learning setups, we commonly face value overestimation issues because the agents can only be trained on a limited set of observable states, historical actions, and historical rewards, without the ability to effectively assess the rewards of OOD actions and corresponding future states.
As a result, inaccurate value estimates for future states can lead to ineffective TD learning, typically causing overly optimistic value estimates due to bootstrapping.
In our formulation of partial-offline RL, the agent can explore different actions within the fixed training set, evaluating new policies with online updated reward feedback.

More specifically, in conventional offline RL formulation, we cannot directly obtain $r_t=R(s_t, s_{t+1})$ from the environment, where $s_{t+1} \sim P(s_{t+1} \mid s_t, a_t)$ and $a_t$ is an out-of-domain action generated by the agent. 
This is because, despite knowing the exact definition of the reward function $R(\cdot)$, accurately estimating $s_{t+1}$ is highly challenging due to the paradox of limited training data and the non-stationary nature of market dynamics.
In partial-offline RL with decoupled state spaces, we instead have $r_t=R(h_t, h_{t+1}, z_t, z_{t+1})$, where $h_{t+1} \sim P_h(h_{t+1} \mid h_t)$ and $z_{t+1} \sim P_z(z_{t+1} \mid z_t, a_t)$.
Here, we still use a pre-defined reward function $R(\cdot)$ and can directly compute the next-step balance state $z_{t+1}$ given an out-of-domain $a_t$.
Although accurately estimating the distribution of the next-step market state $h_{t+1}$ remains intractable, it is independent of $a_t$---\modelname{} leverages this property to approximate the \textit{worst-case} reward $r_t$ in TD learning, using Monte Carlo sampling over carefully designed transformations applied to $h_{t+1}$. 
For further details, refer to the proposed transformation-based TD learning method, which offers a novel solution to mitigating the value overestimation issue.
We will demonstrate in later sections that, due to a better use of the partial-offline properties, the proposed TD method outperforms existing conservative offline RL methods, such as CQL~\cite{kumar2020conservative} and IQL~\cite{kostrikov2021offline}, remarkably in tackling portfolio optimization tasks.

\section{MetaTrader}

In this section, we first review state-of-the-art RL-based methods for portfolio optimization, followed by a detailed presentation of the three key contributions of \modelname{}: (1) the specialized data transformation methods to simulate OOD financial data, (2) a bilevel RL framework that explicitly optimizes both in- and out-of-domain trading performance, and (3) a novel TD learning method that approximates worst-case TD estimates from a batch of transformed TD targets.

\vspace{-5pt}
\paragraph{Revisiting RL-based trading methods.}
\label{sec:stockformer}

We use StockFormer \cite{gao2023stockformer} as an example.
Despite its state-of-the-art performance, a potential drawback lies in the straightforward use of conventional RL methods for offline data.
StockFormer has three network branches $f_{\psi_{1,2,3}}(\cdot)$ to extract the cross-stock relational features $h_t^\text{relat} \in \mathbb{R}^{|S|\times D}$, the long-term predictive features $h_t^\text{long} \in \mathbb{R}^{|S|\times D}$, and the short-term predictive features $h_t^\text{short} \in \mathbb{R}^{|S|\times D}$ from the stock data $o_t=[o_{t-T+1:t}^\text{price}, o_{t-T+1:t}^\text{stat}, o_{t-T+1:t}^\text{cov}]$ in the past $T$ days. $D$ represents the dimension of the hidden features per stock.
The feature extraction module is frozen during policy optimization. These features are used as the input states of the Soft Actor-Critic (SAC) algorithm~\cite{haarnoja2018soft}:
\begin{equation}
    \begin{aligned}
        & \text{Market state encoding:} \quad h_t^\text{relat}=f_{\psi_1}(o_t), \quad 
        h_t^\text{long}=f_{\psi_2}(o_t), \quad 
        h_t^\text{short}=f_{\psi_3}(o_t), \\
        & \text{Actor:} \quad a_t \sim \pi_\theta(h_t^\text{relat}, h_t^\text{long}, h_t^\text{short}, z_t), \quad \text{Critic:} \ q_t \sim Q_\phi(h_t^\text{relat}, h_t^\text{long}, h_t^\text{short}, z_t, a_t), \\
    \end{aligned}
    \label{eq:stockformer}
\end{equation}
where $z_t \in \mathbb{R}^{|S|\times 1}$ represents the holding amount of all trading assets at a certain time step. 
Our approach follows the basic network architectures of StockFormer, including the feature extraction module $f_{\psi_{1,2,3}}$, the actor module $\pi_\theta$, and the critic module $Q_\phi$.

Most existing RL-based portfolio optimization methods train the agent in a manner similar to StockFormer, in which the trading policies are exclusively optimized within a specific offline dataset. 
By maximizing cumulative rewards, this approach carries the risk of overfitting to optimal behaviors in a fixed dataset, potentially leading to ineffective policies when faced with the unobserved dynamics of a non-stationary market in the future.

%
%
%

\begin{figure*}[!t]
    \centering
    \includegraphics[width=\textwidth]{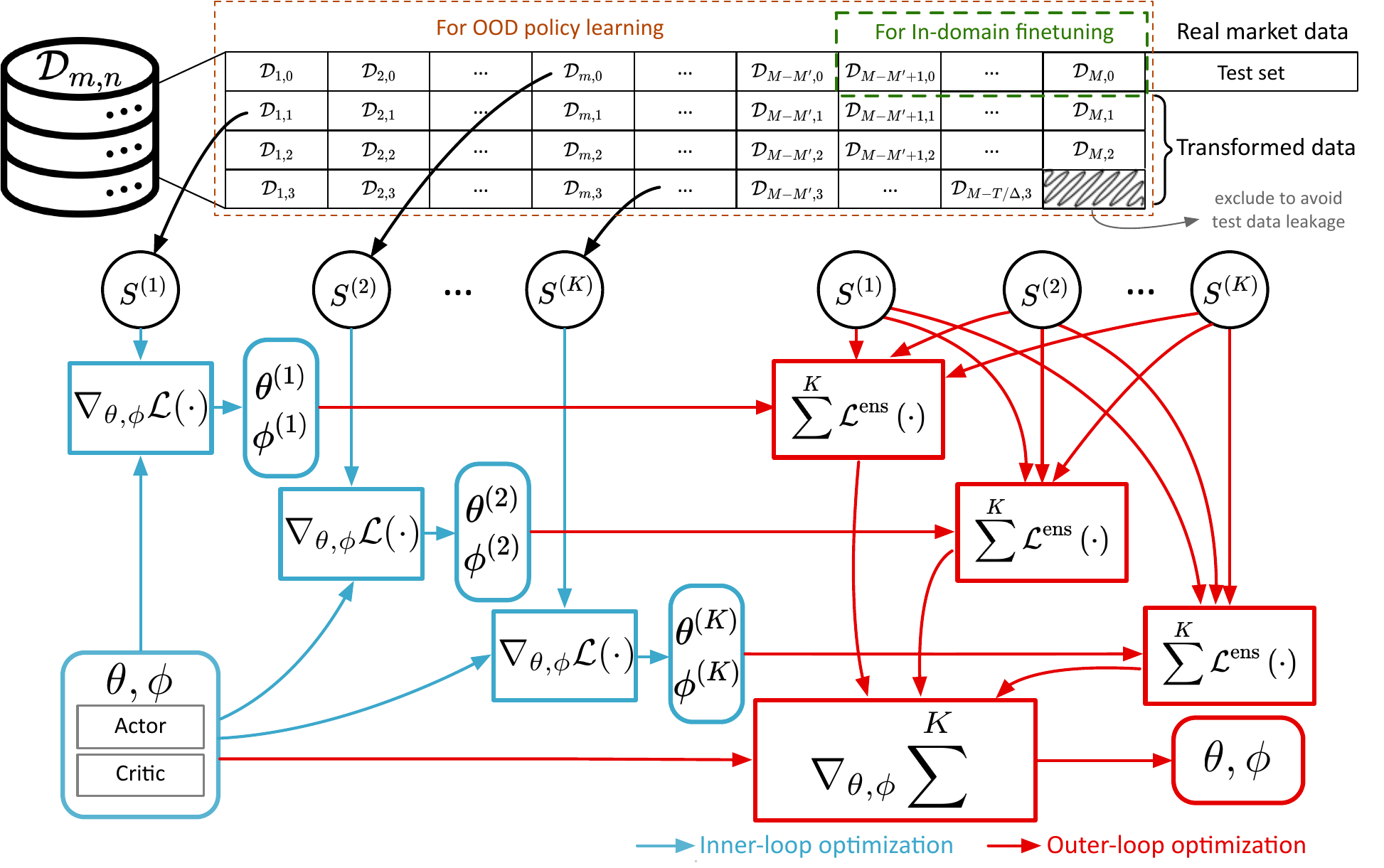}
    \vspace{-20pt}
    \caption{\textbf{The bilevel learning scheme of \modelname{} based on transformed market data.} In the inner optimization loop (\textit{blue arrows}), we optimize the model parameters on a batch of data subsets. In the outer optimization loop (\textit{red arrows}), we perform bilevel gradient updates by explicitly evaluating the inner-loop parameters against another batch of data subsets. This process leads to a more generalizable agent and prevents overfitting to the in-domain optimal policy.
    }
    \label{fig:model}
\end{figure*}




\vspace{-5pt}
\paragraph{Training stages and data transformations.} 
As shown in Figure \ref{fig:model}, we first partition the entire offline training set into multiple subsets, denoted as $\{\mathcal{D}_m\}_{m=1}^{M}$. These subsets are then separated into sequences of $T$ time steps\footnote{In our implementation, we use $T=64$ to approximate the number of trading days in a quarter of the year in the stock market.}. 
Given the non-stationary nature of market data, we employ a two-stage training process: a policy learning stage on the first $M-M^\prime$ subsets, followed by a finetuning stage on the more recent $M^\prime$ subsets.
This is grounded in the fundamental assumption in online temporal data applications: training data closer to the test set may better capture trend patterns that align with those in the test set.

To improve the generalizability of the learned policy to market conditions with notable distribution shifts from the training set, we expand the training subsets by generating a diverse range of OOD market data, using them exclusively in the initial policy learning stage. 
In contrast, we use only the original training data during the finetuning stage to guide the policy toward scenarios more aligned with the test data. 
Accordingly, we name the two training stages in \modelname{} as (1) \textit{OOD policy learning} and (2) \textit{in-domain finetuning}.

To design effective data transformation methods, we treat market data as multivariate time series, whose dynamic patterns can typically be viewed as a combination of three components: short-term randomness, long-term trends, and multi-scale dynamics.
Accordingly, we introduce three data transformation methods $F_{1:3}$ to simulate OOD yet plausible market changes that have not been included in the training set, with each method focusing on one of the three dynamic components:
\begin{itemize}[leftmargin=*] 
    \item $F_1$: At each time step, we select the top $\alpha\%$ assets with the highest price gains and invert their growth rates to simulate unexpected short-term disruptions. 
    \vspace{-3pt}\item $F_2$: We reverse the overall trends in each training subset to simulate the long-term impact of sudden market events, aiming to assess the policy's robustness in such scenarios.
    \vspace{-3pt}\item $F_3$: We downsample the training data by $\Delta$ time steps, squeezing the original temporal dynamics. This enables the model to capture multi-scale temporal correlations with greater flexibility.
\end{itemize}
More technical details are provided in Supplementary Section \textcolor{blue}{S1}.
%
By applying the above transformation methods, we expand the subset collections to $\{\mathcal{D}_{m,n}\}_{m=1,n=0}^{M,N}$ during OOD policy learning, where $n=0$ denotes the original data split.
$N$ is scalable in \modelname{} by adjusting hyperparameters (\textit{e.g.}, $\alpha$, $T$, and $\Delta$) in the data transformation functions. As $N$ increases, we can achieve (1) a broader expansion of the data distribution for policy learning, increasing the likelihood of covering OOD market dynamics, and (2) more accurate Monte Carlo estimations for the expected worst-case future payoffs during TD learning, which we will discuss later. 
For simplicity and without loss of generality, we use $N=3$ with $\alpha=10$, $T=64$, and $\Delta=4$ in our stock trading experiments.

We organize various partitions of the original data along with a diverse range of transformed data into a unified replay buffer as training subsets. We then apply the two-stage bilevel RL scheme across different subsets to enable more generalizable policy optimization.

\begin{figure}[t]
    \centering
    \includegraphics[width=\columnwidth]{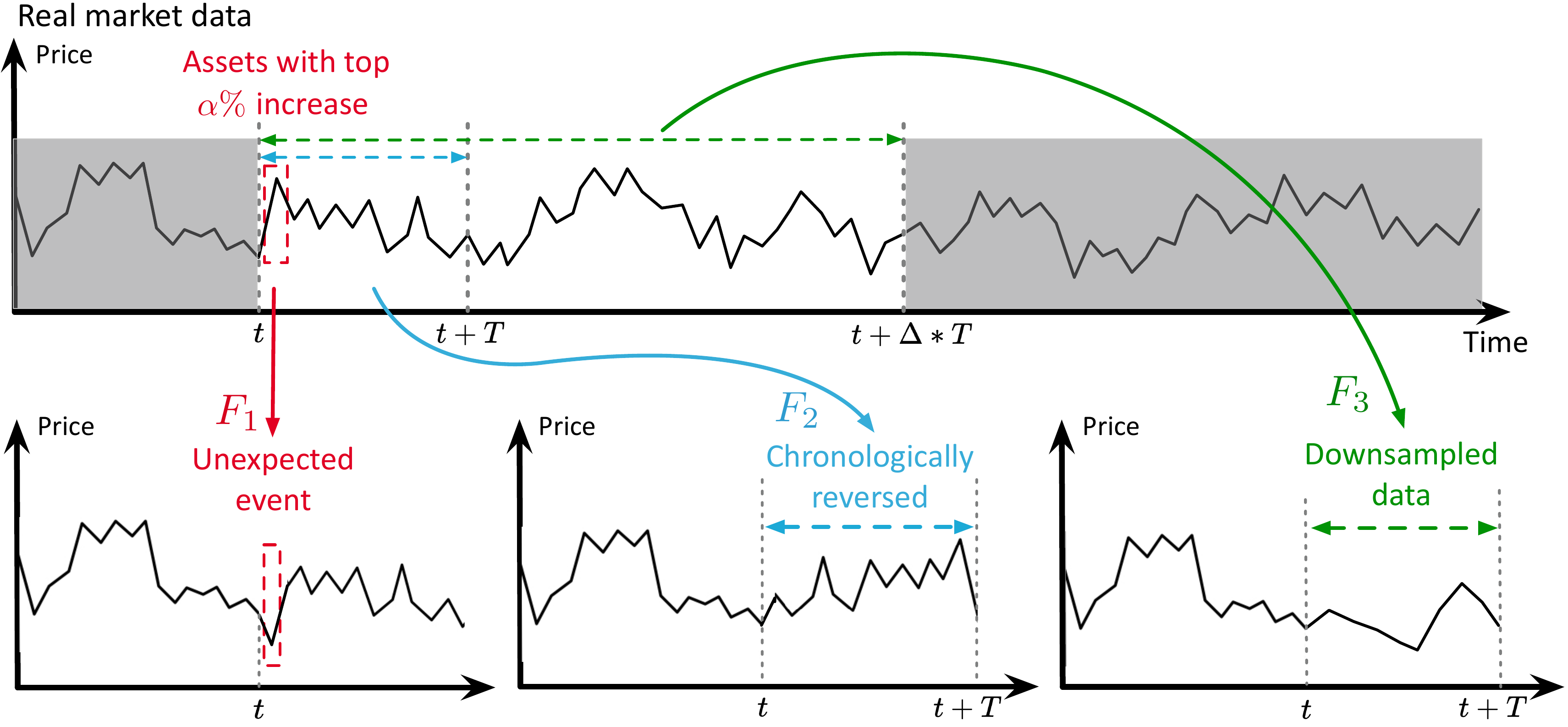}
    \vspace{-15pt}
    \caption{
    \textbf{An example of market data transformations.} 
    $F_1$ selects the top $\alpha\%$ of assets with the highest price gains and inverses their original growth rate to declines to simulate unexpected short-term disruptions. $F_2$ reverses the temporal order of a $T$-length sequence to simulate the long-term impact of certain events. $F_3$ downsamples the original data by $\Delta$ time steps to simulate squeezed global dynamics.
    }
    \label{fig:augment}
\end{figure}


\begin{figure*}[t]
\centering
\begin{minipage}{0.49\textwidth}
   \begin{algorithm*}[H]
      \caption{OOD Policy Learning}
      \label{alg:main}
      \small
      \SetAlgoLined
      \textbf{Input: }{Expanded datasets $\{\mathcal{D}_{m,n}\}_{m=1,n=0}^{M,N}$} \\
      \textbf{Parameters: }{$\alpha_1$, $\alpha_2$, $\eta_1$, $\eta_2$} \\ 
      Randomly initialize $\theta$, $\phi_1$, $\phi_2$ \\
      \DontPrintSemicolon
      \For{$T_1$ steps}{
         Sample $\{\mathcal{D}^{(i)}\}_{i=1}^K \sim \{\mathcal{D}_{m,n}\}_{m=1,n=0}^{M,N}$ \\
         \For{each $\mathcal{D}^{(i)} \in \{\mathcal{D}^{(i)}\}_{i=1}^K$}{
            Sample a batch of data $\mathcal B^{(i)} \sim \mathcal D^{(i)}$ \\
            $\phi_1^{(i)} \leftarrow \phi_1 - \eta_1 \nabla_{\phi_1} \mathcal{L}_Q\big( \mathcal{B}^{(i)}; \ \phi_1 \big)$ \\
            $\phi_2^{(i)} \leftarrow \phi_2 - \eta_1 \nabla_{\phi_2} \mathcal{L}_Q\big( \mathcal{B}^{(i)}; \ \phi_2 \big)$ \\
            $\theta^{(i)} \leftarrow \theta - \alpha_1 \nabla_\theta \mathcal{L}_\pi\big(\mathcal{B}^{(i)}; \ \theta, \phi_1^{(i)}\big)$
         }
         $\phi_1 \leftarrow \phi_1-\eta_2 \textcolor{blue}{\sum_i} \textcolor{red}{\sum_j} \nabla_{\phi_1} \mathcal{L}_Q^\text{ens}\big(\textcolor{blue}{\mathcal{B}^{(i)}}; \ \textcolor{red}{\phi_{1}^{(j)}}\big)$\\
         $\phi_2 \leftarrow \phi_2-\eta_2 \textcolor{blue}{\sum_i} \textcolor{red}{\sum_j} \nabla_{\phi_2} \mathcal{L}_Q^\text{ens}\big(\textcolor{blue}{\mathcal{B}^{(i)}}; \ \textcolor{red}{\phi_{2}^{(j)}}\big)$ \\
         $\theta \leftarrow \theta-\alpha_2 \textcolor{blue}{\sum_i} \textcolor{red}{\sum_j} \nabla_\theta \mathcal{L}_\pi\big(\textcolor{blue}{\mathcal{B}^{(i)}}; \ \textcolor{red}{\theta^{(j)}, \phi_{1}^{(j)}}\big)$
      }
   \end{algorithm*}
\end{minipage}
\hfill
\begin{minipage}{0.5\textwidth}
   \begin{algorithm*}[H]
      \caption{In-Domain Finetuning}
      \label{alg:finetune}
      \small
      \SetAlgoLined
      \textbf{Input: }{Real market data $\{\mathcal{D}_{m,n=0}\}_{m=M-M^\prime+1}^{M}$} \\
      \textbf{Parameters: }{$\alpha_1$, $\alpha_2$, $\eta_1$, $\eta_2$} \\ 
      \DontPrintSemicolon
      Obtain the learned $\theta$, $\phi_1$, $\phi_2$ from Algorithm~\ref{alg:main} \\
      \For{$T_2$ steps}{
         Sample $\{\mathcal{D}^{(i)}\}_{i=1}^{K} \sim
\{\mathcal{D}_{m,n=0}\}_{m=M-M^\prime+1}^{M}$. \\
         \For{each $\mathcal{D}^{(i)} \in \{\mathcal{D}^{(i)}\}_{i=1}^K$}{
            Sample $\mathcal B^{(i)}_\text{tr}, \mathcal B^{(i)}_\text{ts} \sim \mathcal D^{(i)}$ \\
            $\phi_{1}^{(i)} \leftarrow \phi_1 - \eta_1 \nabla_{\phi_1} \mathcal{L}_Q\big(\mathcal{B}^{(i)}_\text{tr}; \ \phi_1 \big)$ \\
            $\phi_2^{(i)} \leftarrow \phi_2 - \eta_1 \nabla_{\phi_2} \mathcal{L}_Q\big(\mathcal{B}^{(i)}_\text{tr}; \ \phi_2 \big)$ \\
            $\theta^{(i)} \leftarrow \theta - \alpha_1 \nabla_\theta \mathcal{L}_\pi\big(\mathcal{B}^{(i)}_\text{tr}; \ \theta, \phi_1^{(i)}\big)$
         }
       $\phi_1 \leftarrow \phi_1-\eta_2 \sum_i \nabla_{\phi_1} \mathcal{L}_Q\big(\mathcal{B}^{(i)}_\text{ts}; \ \phi_1^{(i)}\big)$ \\
       $\phi_2 \leftarrow \phi_2-\eta_2 \sum_i\nabla_{\phi_2} \mathcal{L}_Q\big(\mathcal{B}^{(i)}_\text{ts}; \ \phi_2^{(i)}\big)$ \\
       $\theta \leftarrow \theta-\alpha_2 \sum_i\nabla_\theta \mathcal{L}_\pi\big(\mathcal{B}^{(i)}_\text{ts}; \ \theta^{(i)}, \phi_1^{(i)}\big)$
      }
   \end{algorithm*}
\end{minipage}
\end{figure*}

\vspace{-5pt}
\paragraph{Bilevel learning across transformed data.}

As shown in Algorithm~\ref{alg:main}, we first sample training subsets randomly, $\{\mathcal{D}^{(i)}\}_{i=1}^{K} \sim  \{\mathcal{D}_{m,n}\}_{m=1,n=0}^{M,N}$. 
We then perform an \textit{inner-loop} optimization step to derive $K$ sets of hypothetical model parameters for the actor and critics, denoted by $\theta^{(i)}$ and $\phi_{k}^{(i)}$ for each individual data subset. Here, we employ double Q-networks, parameterized by $\phi_{1,2}$, with $k$ representing their index. 
The objective of inner-loop optimization is to maximize in-domain rewards for each subset.

We proceed with \textit{outer-loop} optimization, as illustrated in Figure \ref{fig:model}, to compute second-order derivatives by evaluating the \textit{inner-loop} parameters $\theta^{(j)}$ and $\phi_{k}^{(j)}$, learned from subset $j$, on the data split $\mathcal{B}^{(i)}$ from distinct subsets.
Notably, this approach distinguishes itself from most existing meta-RL methods by conducting bilevel gradient updates across distinct pairs of subsets.
The aim is to update model parameters to improve the policy's robustness to OOD trajectories.

We formulate the actor's objective function \sloppy $\mathcal{L}_\pi$ as follows, where $s_t=[h_t^\text{relat}, h_t^\text{long}, h_t^\text{short}, z_t]$ and $Z_{\phi_1}$ is a normalization factor:
\begin{equation}
\min_\theta \mathbb{E}_{s_t} \big[D_\mathrm{KL}(\pi_{\theta}(a_t \mid s_t) \ \| \ {\exp(Q_{\phi_1}(s_t,a_t))} / {Z_{\phi_1}(s_t)})\big].
\end{equation}
In subsequent sections, we will elaborate on the inner-loop critic loss $\mathcal{L}_Q$, which minimizes the original TD errors, and the outer-loop critic loss $\mathcal{L}_Q^\text{ens}$, which optimizes the modified ensemble-based TD errors.

\vspace{-5pt}
\paragraph{Bilevel finetuning across in-domain data.}
Due to the non-stationary nature of the time-evolving market, finetuning \modelname{} on recent training data close to the test set can enhance its final performance.
In Algorithm~\ref{alg:finetune}, we employ the bilevel optimization scheme \emph{within} each training subset.
We first draw subsets from the buffer of raw data, such that $
\{\mathcal{D}^{(i)}\}_{i=1}^{K} \sim
\{\mathcal{D}_{m,n=0}\}_{m=M-M^\prime+1}^{M}$. 
Importantly, we exclusively use the original data to eliminate the unexpected noise introduced by the transformed data.
It is essential to note that during the finetuning phase, we perform the inner-loop and outer-loop gradient steps on separate data batches, $\mathcal B^{(i)}_\text{tr}$ and $\mathcal B^{(i)}_\text{ts}$, sampled from the same subset $\mathcal D^{(i)}$.
This approach aims to facilitate model adaptation to recent market dynamics.

\begin{figure}[t]
    \centering
    \includegraphics[width=0.95\columnwidth]{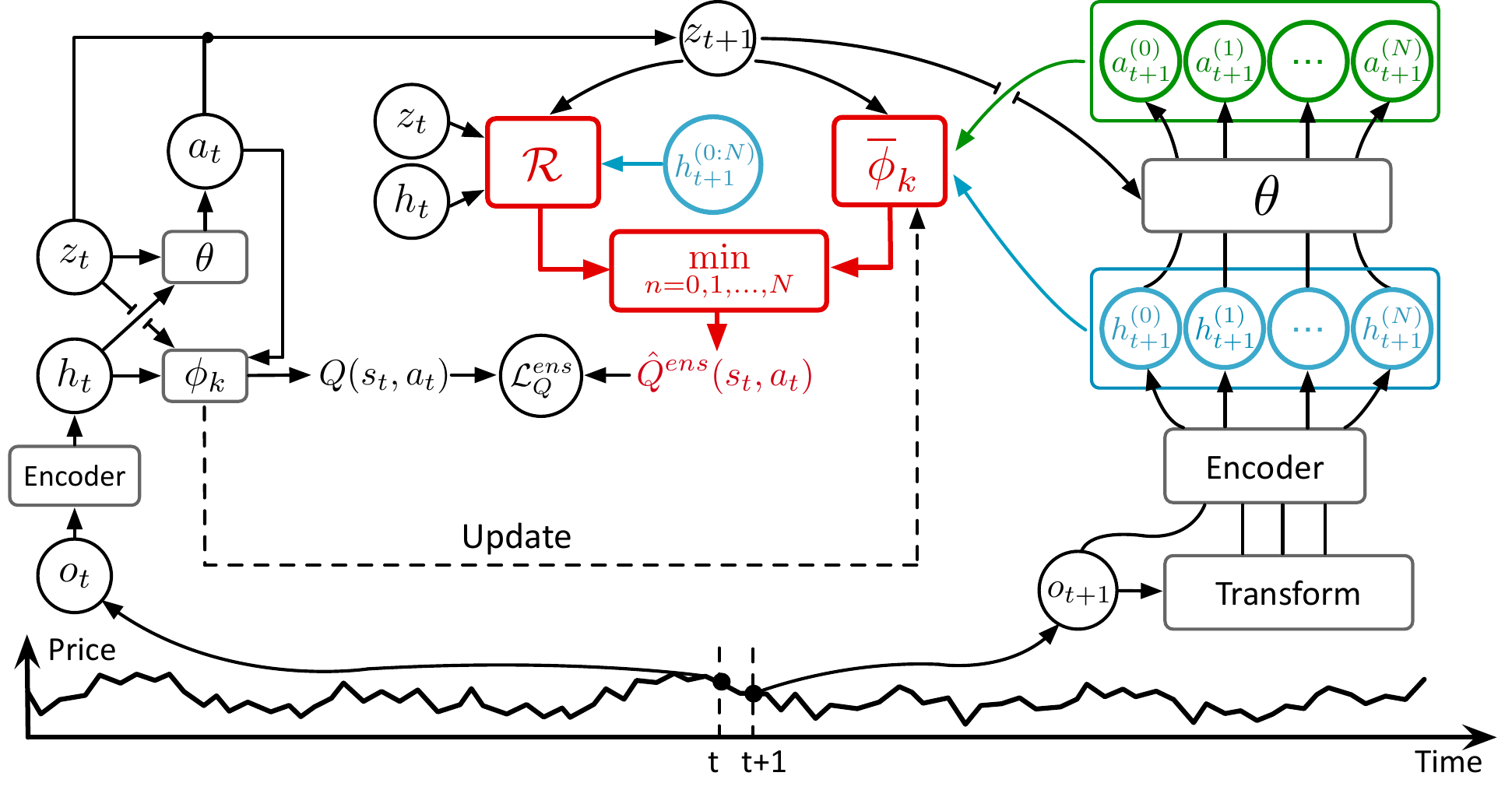}
    \vspace{-10pt}
    \caption{\textbf{Transformation-based TD learning with worst-case bootstrapping.}
    The left part represents the TD estimate, and the right part corresponds to the TD target. By approximating worst-case future payoffs through a Monte Carlo method over a batch of data transformations, our approach aims to improve the generalizability of policies learned from offline data. This also helps mitigate the value overestimation issue, which is especially problematic when there are substantial discrepancies between the training and test distributions of non-stationary market data.
    }
    \label{fig:TD_overview}
\end{figure}

\vspace{-5pt}
\paragraph{Transformation-based TD learning with worst-case bootstrapping.}
\label{sec:TD}
We propose a novel TD method for training the critic model during the OOD policy learning phase.
In Algorithm~\ref{alg:main}, the training objectives of $Q_{\phi_{1,2}}$, including the inner-loop $\mathcal{L}_Q$ and the outer-loop $\mathcal{L}_Q^\text{ens}$, can be formulated as 
\begin{equation}
\label{eq:q_function}
    \min_{\phi_k}\mathbb{E}_{(s_t,a_t)}\big[Q_{\phi_k}(s_t,a_t)-\mathrm{sg}\big(\widehat{Q}(s_t,a_t)\big)\big]^2,
\end{equation}
where $Q_{\phi_k}(\cdot)$ represents the TD estimate of the critic $k$ at timestamp $t$, $\widehat{Q}(\cdot)$ represents the corresponding TD target, and $\mathrm{sg}(\cdot)$ denotes stopping the gradient backpropagation.
We here denote $s_t = [h_t, z_t]$ and $h_t=[h_t^\text{relat}, h_t^\text{long}, h_t^\text{short}]$.
In the inner-loop optimization step, we formulate the TD target $\widehat{Q}(\cdot)$ as 
\begin{equation}
\label{eq:q_target_in}
    \widehat{Q}(s_t, a_t)  =  R(s_t, s_{t+1}) +\gamma\big[
    -\lambda \log \pi_{\theta} ({a}_{t+1} \mid {s}_{t+1})  
    + \min_{k=1,2} {Q_{\bar{\phi}_k}({s}_{t+1}, {a}_{t+1})} \big],
\end{equation}
where $R(\cdot)$ is the pre-defined reward function and ${a}_{t+1}$ is generated by the policy $\pi_{\theta}\left(\cdot \mid s_{t+1}\right)$. 
We incorporate double target Q-networks $Q_{\bar{\phi}_{1,2}}$, which are updated using the moving-average parameters from corresponding Q-networks $Q_{\phi_{1,2}}$.
$Q_{\bar{\phi}_k}$ is the next-step Q-value from each target Q-network.

In the outer-loop gradient update step, as shown in Figure \ref{fig:TD_overview}, we incorporate a new form of TD target derived from a batch of transformed data in Eq.~\eqref{eq:q_function}. For clarity, we denote $\widehat{Q}(\cdot)$ as
\begin{equation}
\begin{aligned}
\label{eq:q_target_out}
    \widehat{Q}^\text{ens} (s_t, a_t) = \textcolor{blue}{\min_{n=0:N}} \Big[R(s_t, s_{t+1}^{(n)}) +\gamma \big( 
    -\lambda \log \pi_{\theta} ({a}_{t+1}^{(n)} \mid {s}_{t+1}^{(n)})
    + \min_{k=1,2} \textcolor{blue}{Q_{\bar{\phi}_k}({s}_{t+1}^{(n)}, {a}_{t+1}^{(n)})} \big) \Big],
\end{aligned}
\end{equation}
where $\{{s}_{t+1}^{(n)}\}_{n=1}^N$ represent simulated next-step market states transformed by $F_{1:3}$, and ${a}_{t+1}^{(n)}$ is generated by the policy $\pi_{\theta}(\cdot \mid s^{(n)}_{t+1})$. Notably, $s_{t+1}$ is specifically referred to as ${s}_{t+1}^{(0)}$, denoting the next-step market state encoded from the original data.

Eq.~\eqref{eq:q_target_out} is feasible in our partial-offline RL formulation, where $h_{t+1} \sim P_h(h_{t+1} \mid h_t)$ and $z_{t+1} \sim P_z(z_{t+1} \mid z_t, a_t)$. 
This decoupled state transition approach allows us to directly evaluate the value of $a_t$ given $h_t$ and $z_t$, since in $R(h_t, h_{t+1}, z_t, z_{t+1})$, the only missing component is the next market state $h_{t+1}$, which is independent of the action.
As previously described, while accurately estimating the distribution of $h_{t+1}$ is intractable, we perform Monte Carlo sampling over diverse transformations of the original data at $t+1$ and approximate the \textit{worst-case} TD target using $\{h_{t+1}^{(n)}\}_{n=1}^N$.

Furthermore, it is important to note that existing ensemble-based TD methods~\cite{an2021uncertainty,lee2022offline,wu2021aggressive} typically train multiple target Q-networks with separate model parameters and compute ensemble value regularization by exploiting the implicit diversity among these Q-networks. 
In contrast, our approach uses a single pair of target Q-networks and derives the worst-case TD target by leveraging the explicit diversity introduced by transformed data.

\section{Results}
\label{sec:exp}

\begin{table*}[t]
\vspace{-10pt}
\footnotesize
	\centering
         \caption{
        \textbf{Offline evaluation results on CSI and NASDAQ datasets}. 
        We use \textit{cumulative return}, \textit{annualized return}, \textit{Sharpe ratio}, and \textit{maximum drawdown} as the metrics. 
        Given the inherent instability of RL algorithms, we present the results of RL-based models from $10$ random training seeds.
        }
	\vspace{5pt}
	\setlength\tabcolsep{3pt}
	\begin{tabular}{l|rrrr|rrrr}
		\toprule
		\multirow{2}{*}{Method} & \multicolumn{4}{c|}{CSI-300} & \multicolumn{4}{c}{NASDAQ-100} \\
	     & CR$^\uparrow$ & {AR$^\uparrow$} & SR$^\uparrow$ &MDD$^\downarrow$& CR$^\uparrow$ & {AR$^\uparrow$} & SR$^\uparrow$ &MDD$^\downarrow$\\
      \midrule
	     Market benchmark &0.08 & 0.02 &  0.23 &0.31& 0.99 & 0.26 & 0.98&0.28 \\
	    HATR
     & $-$0.05&$-$0.02 &0.06&0.51 &0.10 &0.03 &0.25 &0.35\\
	     Relational Ranking
      &$-$0.13 &$-$0.05  &$-$0.05 &0.37 & 0.79&0.22 &0.75 &0.37 \\
	    AutoFormer
     & $-$0.08& $-$0.03 &0.02&0.58  & $-$0.28& $-$0.10& $-$0.27&0.41 \\
	    FactorVAE
     &0.96 &0.25  &1.25&\textbf{0.17}  &0.90 &0.24 &0.77&\textbf{0.26} \\
	     \midrule
      FinRL-SAC
      &0.83$\pm$\scriptsize\text{0.05} & 0.22$\pm$\scriptsize\text{0.01} & 0.92$\pm$\scriptsize\text{0.04}&0.30$\pm$\scriptsize\text{0.01}&0.37$\pm$\scriptsize\text{0.05}&0.11$\pm$\scriptsize\text{0.01} &0.54$\pm$\scriptsize\text{0.04}&0.32$\pm$\scriptsize\text{0.01} \\
	     FinRL-DDPG
      &0.58$\pm$\scriptsize\text{0.15} &0.16$\pm$\scriptsize\text{0.04}  & 0.73$\pm$\scriptsize\text{0.12}& 0.34$\pm$\scriptsize\text{0.03}&0.91$\pm$\scriptsize\text{0.11} &0.24$\pm$\scriptsize\text{0.02}&0.75$\pm$\scriptsize\text{0.05} &0.41$\pm$\scriptsize\text{0.01}\\
      CQL
      &0.64$\pm$\scriptsize\text{0.07} & 0.18$\pm$\scriptsize\text{0.02} & 0.75$\pm$\scriptsize\text{0.05}&0.33$\pm$\scriptsize\text{0.02} &0.77$\pm$\scriptsize\text{0.12} &0.21$\pm$\scriptsize\text{0.02} &0.76$\pm$\scriptsize\text{0.06}&0.35$\pm$\scriptsize\text{0.02} \\
      IQL
      &1.02$\pm$\scriptsize\text{0.10} & 0.26$\pm$\scriptsize\text{0.02} & 0.94$\pm$\scriptsize\text{0.06}&0.32$\pm$\scriptsize\text{0.02}&0.92$\pm$\scriptsize\text{0.09}&0.24$\pm$\scriptsize\text{0.02} &0.87$\pm$\scriptsize\text{0.04}&0.36$\pm$\scriptsize\text{0.01} \\
      SARL
      & 1.06$\pm$\scriptsize\text{0.14} & 0.27$\pm$\scriptsize\text{0.03} & 0.98$\pm$\scriptsize\text{0.08} & 0.36$\pm$\scriptsize\text{0.02}&1.03$\pm$\scriptsize\text{0.20}& 0.27$\pm$\scriptsize\text{0.04} &0.80$\pm$\scriptsize\text{0.09}&0.40$\pm$\scriptsize\text{0.01}\\
          StockFormer
          & {1.24$\pm$\scriptsize\text{0.10}} &  {0.31$\pm$\scriptsize\text{0.02}}& {1.20$\pm$\scriptsize\text{0.06}}&0.31$\pm$\scriptsize\text{0.02}& 0.98$\pm$\scriptsize\text{0.07}& 0.26$\pm$\scriptsize\text{0.02}& {0.93$\pm$\scriptsize\text{0.04}}&0.32$\pm$\scriptsize\text{0.02} \\
          \midrule
	     \modelname{} & \textbf{1.44$\pm$\scriptsize\text{0.07}} & \textbf{0.35$\pm$\scriptsize\text{0.02}} & \textbf{1.35$\pm$\scriptsize\text{0.08}} & 0.28$\pm$\scriptsize\text{0.02}&
         \textbf{1.30$\pm$\scriptsize\text{0.08}} &\textbf{0.32$\pm$\scriptsize\text{0.02}} &\textbf{1.11$\pm$\scriptsize\text{0.04}} &0.31$\pm$\scriptsize\text{0.00}\\
		\bottomrule
	\end{tabular}
	\label{tab:Quantitative_comparison_offline}
\end{table*}


We evaluate \modelname{} using the CSI-300 and NASDAQ-100 datasets, both adopted from StockFormer~\cite{gao2023stockformer}.
The CSI dataset is sourced from the CSI-300 Composite Index, which includes $88$ stocks. It spans from 01/17/2011 to 04/01/2022 and is split into training and test sets containing $1{,}936$ and $785$ trading days, respectively. 
The NASDAQ dataset contains $86$ NASDAQ stocks, collected from Yahoo Finance. It covers the period from 01/17/2011 to 04/01/2022, with a training set of $2{,}002$ trading days and a test set of $819$ trading days. 
We provide details on data preprocessing, normalization, and the technical indicators used in our method in Supplementary Section \textcolor{blue}{S2}.

We compare \modelname{} with the following models: 
(A) \textit{\textbf{Market benchmarks}}, including the CSI-300 Index and the NASDAQ-100 Index.
(B) \textit{\textbf{RL trading methods}}, including FinRL~\cite{liu2021finrl}, SARL~\cite{ye2020reinforcement}, and StockFormer~\cite{gao2023stockformer}.
(C) \textit{\textbf{Offline RL methods}}, including CQL~\cite{kumar2020conservative} and IQL~\cite{kostrikov2021offline}.
(D) \textit{\textbf{Stock prediction or general time series forecasting methods}}, including HATR~\cite{wang2021hierarchical}, Relational Ranking~\cite{feng2019temporal}, AutoFormer~\cite{wu2021autoformer}, and FactorVAE~\cite{duan2022factorvae}. 
For the stock prediction methods, we apply the \textit{buy-and-hold} strategy, \textit{i.e.}, buying the stock with the highest predicted return over the next $5$ days and selling it $5$ days later.

All models are tested with market transaction costs. Unless otherwise specified, the results for the RL-based methods are averaged across three random training seeds.
For the details on the training hyperparameters, please refer to Supplementary Section \textcolor{blue}{S3}.

\begin{figure*}[!t]
    \centering
    \subfigure[CSI-300]{\includegraphics[width=0.48\textwidth]{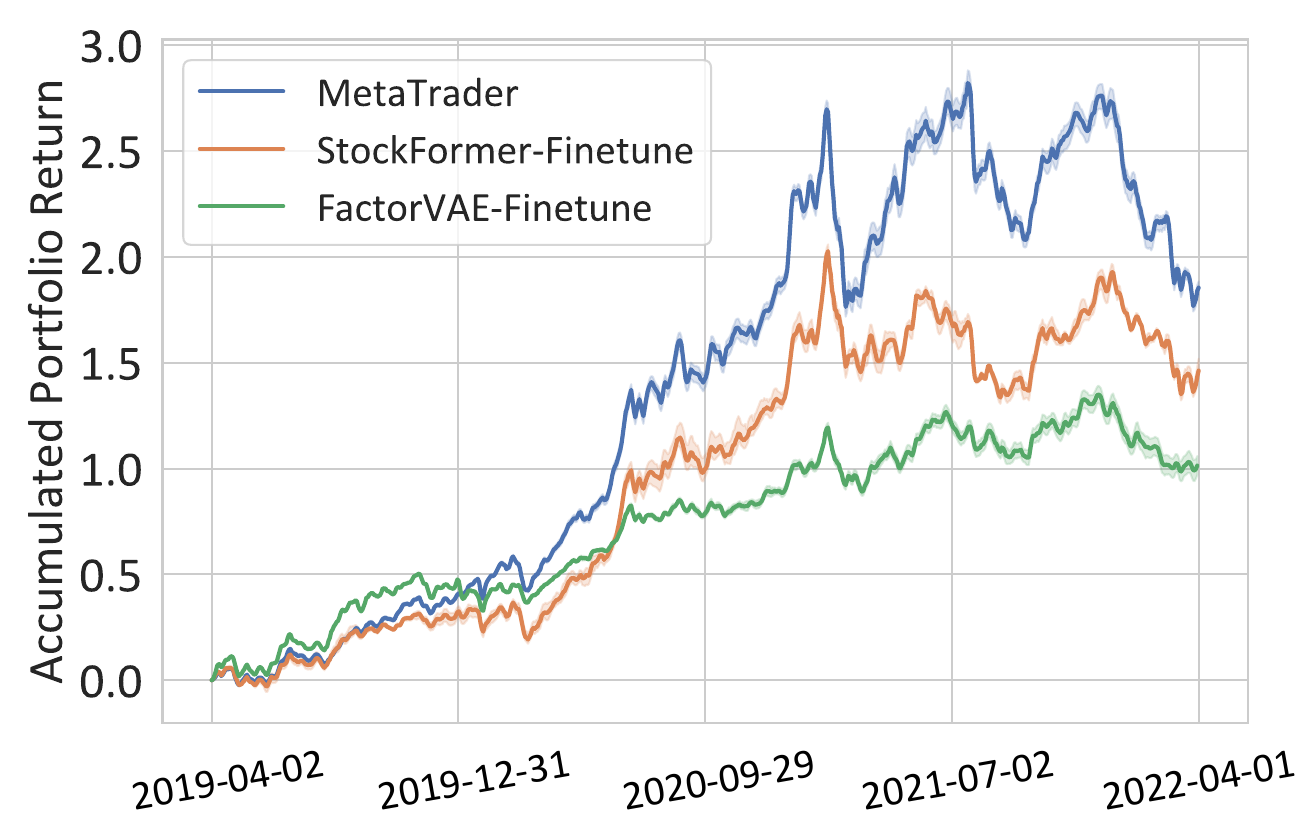}
    \label{fig:visual_testset_csi}
    }
    \subfigure[NASDAQ-100]{\includegraphics[width=0.48\textwidth]{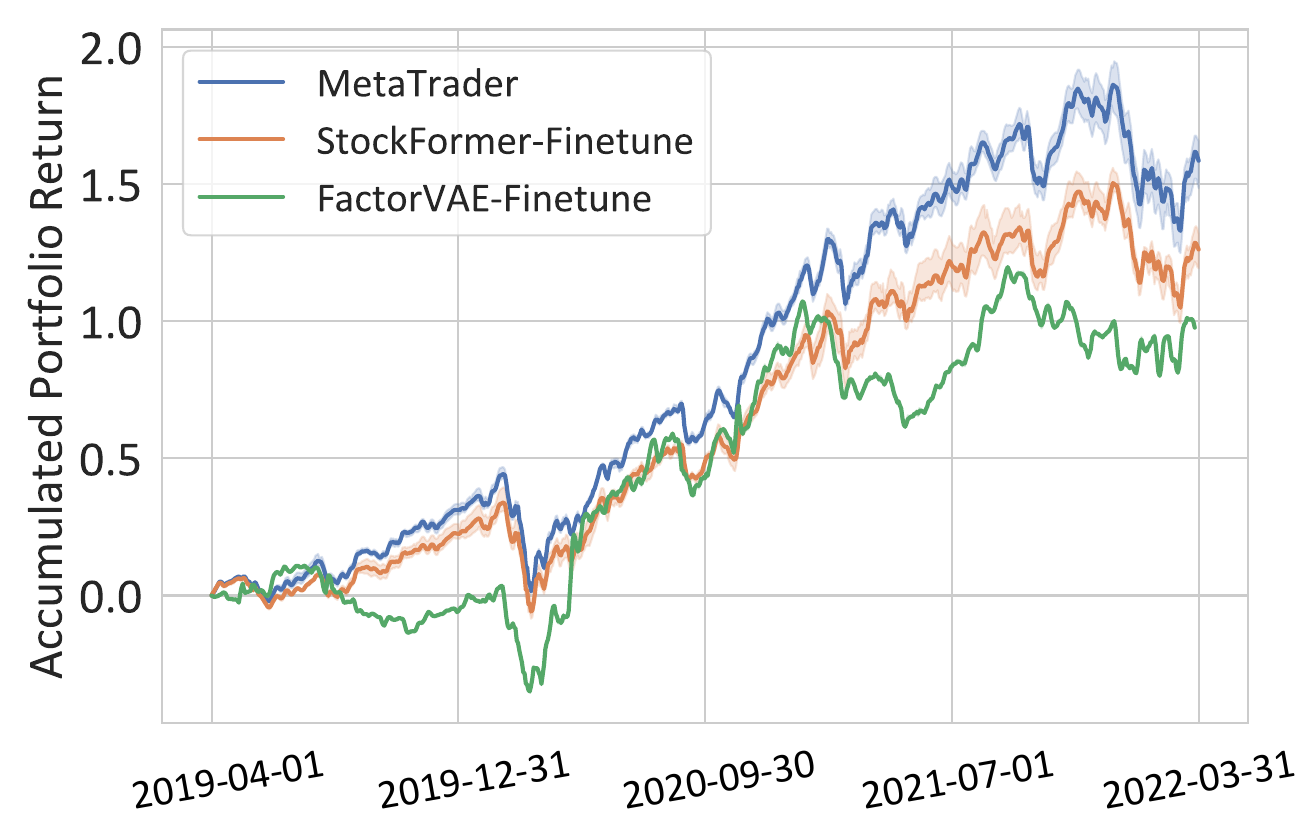}
    \label{fig:visual_testset_nasdaq}
    } 
    \vspace{-10pt}
    \caption{
    \textbf{The cumulative returns under the online adaptation setup.} We divide the entire test set into three equal-length splits and progressively finetune the models over the streaming data. All results are obtained from models trained with $10$ random seeds.
    }
    \label{fig:visual_testset}
\end{figure*}

\begin{figure}[!t]
\begin{center}
\centerline{\includegraphics[width=\columnwidth]{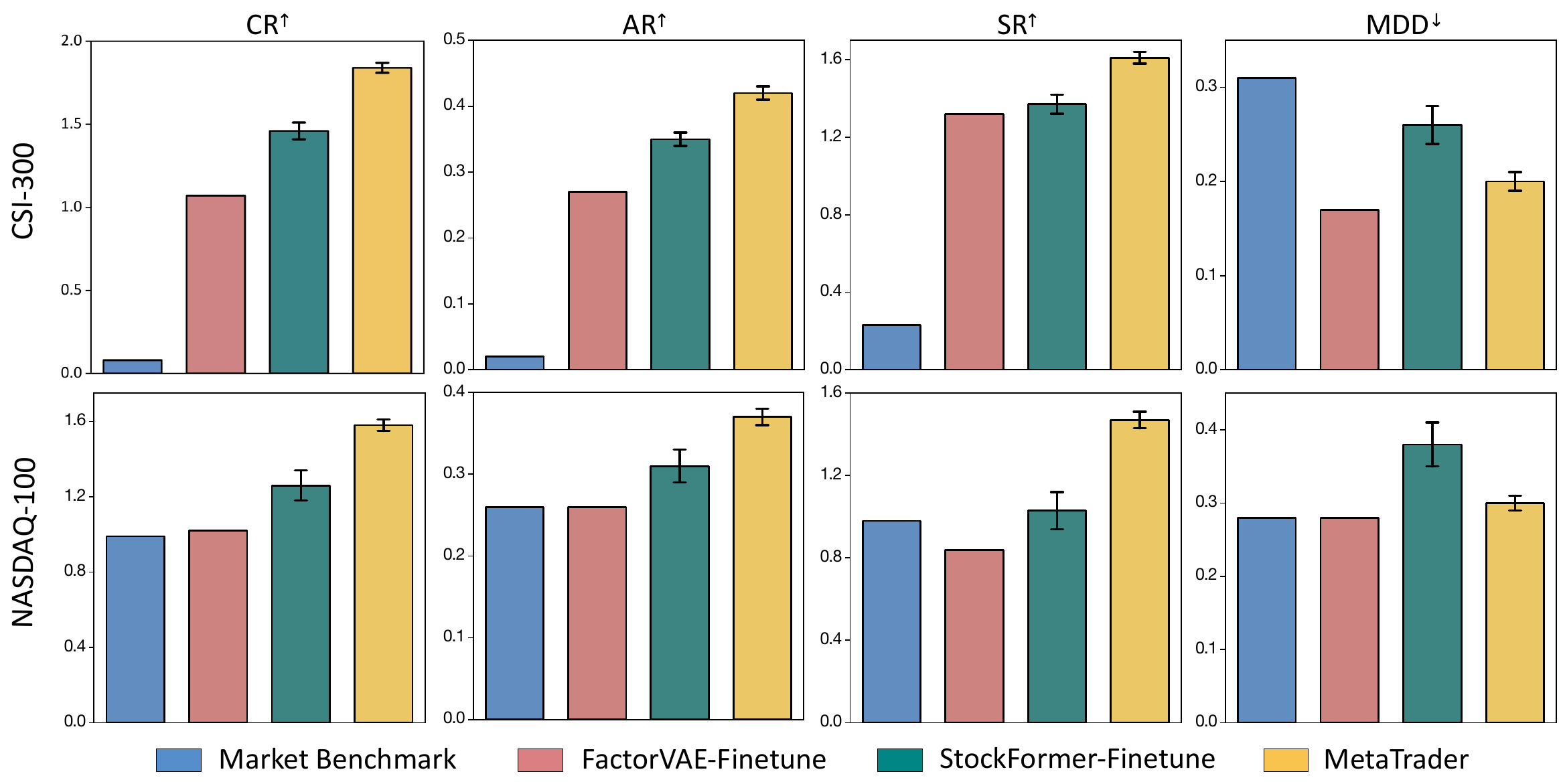}}
\vspace{-10pt}
\caption{\textbf{Full comparisons in all metrics under the online adaptation setup.} The online adaptation setup more effectively demonstrates the advantages of bilevel policy learning and finetuning for efficient domain adaptation, enabling \modelname{} to outperform StockFormer by significant margins.
}
\label{fig:Quantitative_comparison_online}
\end{center}
\vspace{-20pt}
\end{figure}

\begin{figure}[t]
\begin{center}
\centerline{\includegraphics[width=\columnwidth]{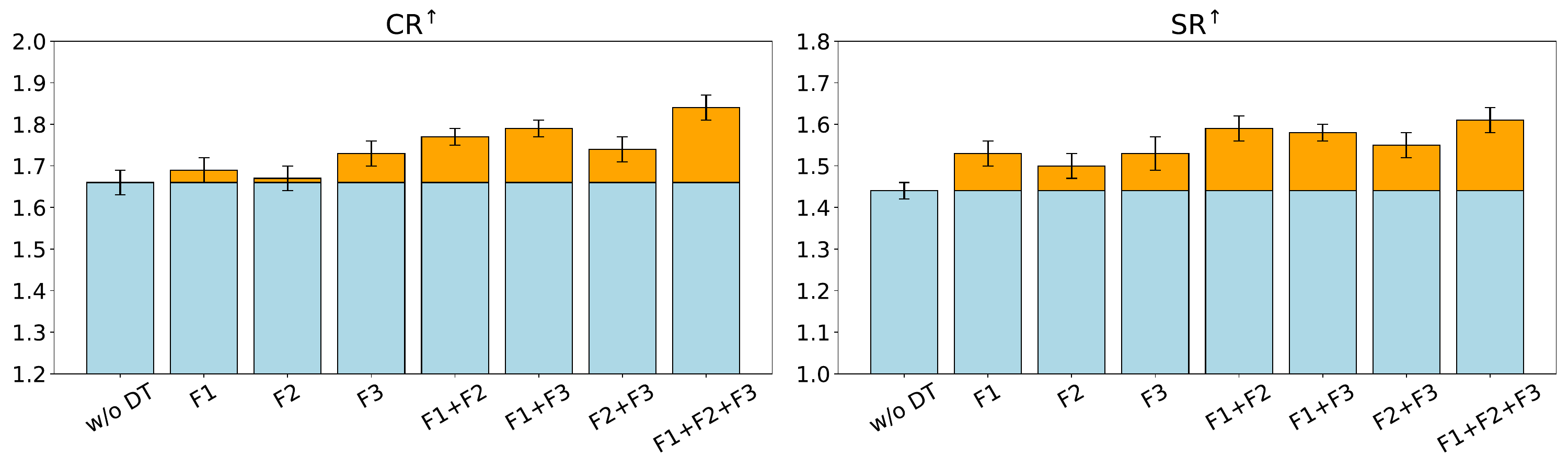}}
\vspace{-15pt}
\caption{\textbf{Analyses of data transformation techniques for \textit{OOD policy learning} (Algorithm~\ref{alg:main}).} We report the mean results on the CSI dataset over $3$ seeds. DT: Data Transformation.
}
\label{fig:ablation_augmentation}
\end{center}
\vspace{-20pt}
\end{figure}

\vspace{-5pt}
\paragraph{Standard offline evaluation.}

For both datasets, we perform OOD policy learning using training data from 01/17/2011 to 12/31/2018. Next, we conduct in-domain finetuning on the last-year training data, specifically from 01/04/2018 to 12/31/2018. 
To ensure no overlap between the test and training sets, we set the input data of the test sequences to start from January 2019, covering the trading days from 04/01/2019 to 04/01/2022 over three years.

Table \ref{tab:Quantitative_comparison_offline} presents the quantitative results of \modelname{} 
in terms of \textit{cumulative return} (CR), \textit{annualized return} (AR), \textit{Sharpe ratio} (SR), and \textit{maximum drawdown} (MDD). Please refer to Supplementary Section \textcolor{blue}{S4}
for detailed definitions of the evaluation metrics. 
It is worth noting that our approach generally outperforms all stock prediction methods by substantial margins.
In particular, compared to FactorVAE, \modelname{} outperforms by $\textbf{50\%}$ in cumulative return ($1.44$ vs. $0.96$) on the CSI dataset and by $\textbf{44.4\%}$ on the NASDAQ dataset ($1.30$ vs. $0.90$).
In finance, the Sharpe ratio (also known as the reward-to-variability ratio) measures the additional amount of return that an investor receives per unit of increase in risk. It is defined as the difference between the returns of the investment and the risk-free return, divided by the standard deviation of the investment returns.
\modelname{} outperforms FactorVAE by $\textbf{44.1\%}$ in Sharpe ratio ($1.11$ vs. $0.77$) on the NASDAQ dataset, demonstrating a strong balance between portfolio returns and risk control.

When compared to other RL-based trading methods, \modelname{} delivers the best performance across all evaluation metrics. It improves upon the state-of-the-art StockFormer method in terms of cumulative return by $\textbf{16.1\%}$ on the CSI dataset and by $\textbf{32.7\%}$ on NASDAQ.
Furthermore, we implement baseline models using the same neural network architecture as \modelname{}, but trained with other conservative offline RL techniques, including CQL and IQL.
As observed, existing offline RL approaches struggle with RL-for-finance tasks due to fluctuations in data distributions, leading to significant shifts between the training and testing domains. 
In contrast, our approach achieves notable performance gains through bilevel policy learning, which effectively prevents the policy from overfitting to the offline data.

\vspace{-5pt}
\paragraph{Online adaptation on streaming data.}

We employ another experimental setup that closely aligns with the dynamic financial decision-making applications, where we finetune the model on-the-fly over the streaming test data. 
The entire test set is divided into three equal-length periods: 04/01/2019---04/01/2020, 04/02/2020---04/01/2021, and 04/02/2021---04/01/2022.
Each period is followed by an in-domain finetuning phase before testing. For instance, for the test period of 04/02/2020---04/01/2021, we perform in-domain finetuning using data in 04/01/2019---04/01/2020 prior to testing.

For the online adaptation setup, our main comparison is between \modelname{}, \textit{FactorVAE-Finetune}, and \textit{StockFormer-Finetune}, all of which are continuously finetuned using the streaming test data.
The results are shown in Figure
\ref{fig:visual_testset} and Figure \ref{fig:Quantitative_comparison_online}.
As we can see, \modelname{} presents a remarkable advantage against other approaches, including the state-of-the-art stock prediction model (\textit{i.e.}, FactorVAE) and RL-based stock trading method (\textit{i.e.}, StockFormer).
On the CSI dataset, it improves StockFormer-Finetune by $\textbf{26\%}$ in cumulative return ($1.84$ vs. $1.46$) and by around $\textbf{18\%}$ in Sharpe ratio ($1.61$ vs. $1.37$).
On the NASDAQ dataset, \modelname{} improves StockFormer-Finetune by over $\textbf{25\%}$ in cumulative return ($1.58$ vs. $1.26$) and by around $\textbf{43\%}$ in Sharpe ratio ($1.47$ vs. $1.03$).


\begin{figure}[t]
\begin{center}
\centerline{\includegraphics[width=\columnwidth]{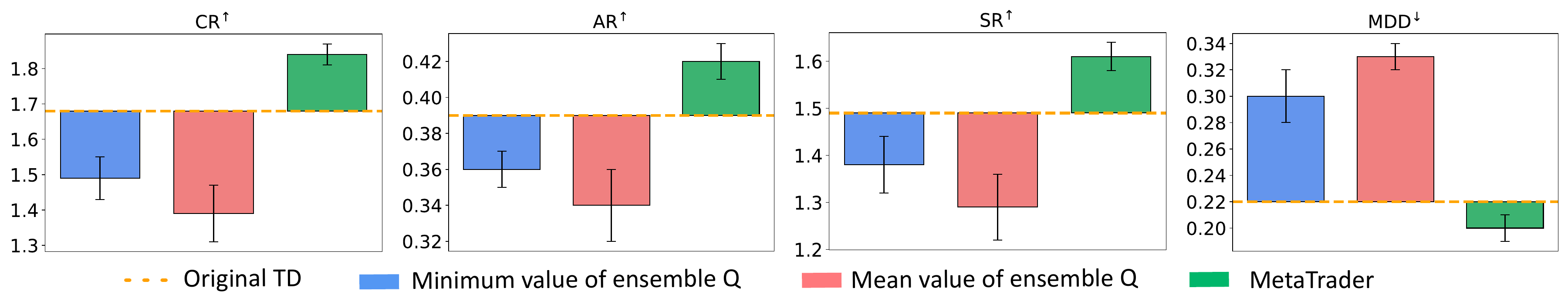}}
\vspace{-10pt}
\caption{\textbf{Ablation studies of transformation-based TD ensembles.} We compare our approach with the following: (1) the use of the original TD method in bilevel learning, and (2-3) baseline models that compute TD targets based on real future market data using an ensemble of target Q-networks~\cite{an2021uncertainty,lee2022offline}. In contrast, during the OOD policy learning stage, \modelname{} computes the TD target using transformed data and a single pair of target Q-networks. The experiments are conducted under the online adaptation setup on the CSI dataset.
}
\label{fig:ablation_TD}
\end{center}
\vspace{-20pt}
\end{figure}

\vspace{-5pt}
\paragraph{The effectiveness of data transformation.}

To assess the true impact of different data transformation techniques, we experiment with baseline models that (1) do not incorporate transformed data at any stage of training, and (2) incorporate only some of the data transformation techniques.
We have two observations from Figure \ref{fig:ablation_augmentation}.
First, leveraging any of the data transformation methods during the OOD policy learning phase consistently enhances the model's final performance, resulting in substantial improvements across all three evaluation metrics.
Second, combining multiple transformation techniques leads to further significant gains. Notably, we observe a $\textbf{10.8\%}$ increase in cumulative return ($1.66\rightarrow 1.84$) for online adaptation on the CSI dataset.

\begin{figure}[t]
\begin{center}
\centerline{\includegraphics[width=0.98\columnwidth]{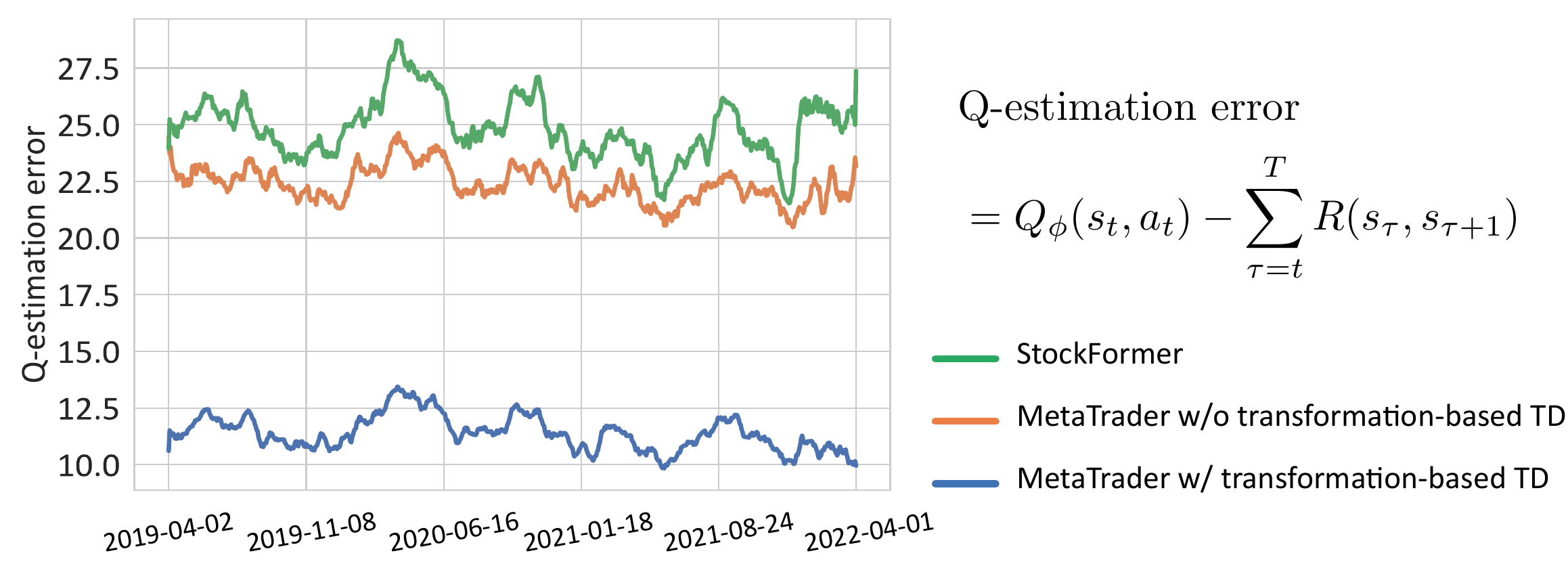}}
\vspace{-10pt}
\caption{\textbf{The disparities between the predicted values by the critic and the true discounted future rewards.} A larger disparity signifies a more pronounced \textit{value overestimation} issue in offline RL. 
The results are obtained under the offline evaluation setup on the CSI dataset. 
}
\label{fig:value-estimation}
\end{center}
\vspace{-20pt}
\end{figure}

\begin{table}[t]
\centering
\caption{\textbf{Ablation studies of the operations in the \textit{in-domain finetuning} stage (Algorithm~\ref{alg:finetune}).} 
We evaluate alternative configurations of the bilevel gradient update and transformed stock data. The experiments are conducted within the online adaptation setup.}
\vspace{5pt}
\small
\setlength\tabcolsep{2pt}
\begin{tabular}{cc|cccc|cccc}
\toprule
 Bilevel &  Transf.  & \multicolumn{4}{c|}{CSI} & \multicolumn{4}{c}{NASDAQ} \\
	     gradient &  data &  {CR}$^\uparrow$ & {AR$^\uparrow$} & SR$^\uparrow$ & {MDD}$^\downarrow$ & {CR}$^\uparrow$ & {AR$^\uparrow$} & SR$^\uparrow$ & {MDD}$^\downarrow$ \\
\midrule
\xmark & \xmark & 1.78$\pm$\scriptsize\text{0.03}&0.41$\pm$\scriptsize\text{0.01}&1.57$\pm$\scriptsize\text{0.03} &0.23$\pm$\scriptsize\text{0.01}&1.41$\pm$\scriptsize\text{0.04}&0.34$\pm$\scriptsize\text{0.01}&1.34$\pm$\scriptsize\text{0.04}&0.31$\pm$\scriptsize\text{0.01} \\
\cmark & \xmark  &\textbf{1.84$\pm$\scriptsize\text{0.03}}&\textbf{0.42$\pm$\scriptsize\text{0.01}}&\textbf{1.61$\pm$\scriptsize\text{0.03}}&\textbf{0.20}$\pm$\scriptsize\text{0.01} &\textbf{1.58$\pm$\scriptsize\text{0.03}} &\textbf{0.37$\pm$\scriptsize\text{0.01}}&\textbf{1.47$\pm$\scriptsize\text{0.04}}&\textbf{0.30$\pm$\scriptsize\text{0.01}} \\
\cmark & \cmark &0.84$\pm$\scriptsize\text{0.04} &0.23$\pm$\scriptsize\text{0.01} &0.94$\pm$\scriptsize\text{0.05} &0.33$\pm$\scriptsize\text{0.02} &1.24$\pm$\scriptsize\text{0.03} &0.31$\pm$\scriptsize\text{0.01} &1.03$\pm$\scriptsize\text{0.05}&0.33$\pm$\scriptsize\text{0.01}\\
\bottomrule
\end{tabular}
\label{tab:ablation_bilevel_and_fn}
\end{table}

\vspace{-5pt}
\paragraph{Impact of the transformation-based conservative TD ensembles.}

To assess the effectiveness of the transformation-based TD method used in the proposed bilevel RL framework, we implement a baseline model that adopts the original TD method from SAC. 
Figure \ref{fig:ablation_TD} demonstrates the improvements achieved by our proposed TD method, with a significant increase of $\textbf{9.5\%}$ in {cumulative} return on the CSI dataset.
Furthermore, we experiment with other TD ensemble approaches~\cite{an2021uncertainty,lee2022offline}, $i.e.$, such as using the minimum and mean values from $5$ parallel Q-networks as the Bellman target. 
As shown in Figure \ref{fig:ablation_TD}, our approach, which computes TD targets using transformed data and a single pair of target Q-networks, demonstrates a significant advantage over other ensemble-based TD learning alternatives.
These results highlight the effectiveness of using diverse future data transformations to approximate worst-case future returns, guiding the agent towards safer and more reliable trading behaviors.

Additionally, in Figure \ref{fig:value-estimation}, we compare the value estimation accuracy with vs. without the transformation-based TD ensembles.
Specifically, we report the discrepancies between the values predicted by the critic models and true values, determined by the discounted sum of rewards throughout the same data trajectories.
As observed, StockFormer and ``\textit{\modelname{} w/ original TD}'' tend to overestimate the true value function.
In contrast, the values estimated by the final ``\textit{\modelname{} w/ ensemble-based TD}'' are notably more accurate and more akin to the true values.

\vspace{-5pt}
\paragraph{Technical designs of the in-domain finetuning stage.}
We perform model finetuning on real data from the most recent year, using bilevel gradient updates (see Algorithm~\ref{alg:finetune}).
%
%
In Table \ref{tab:ablation_bilevel_and_fn}, we investigate the necessity of bilevel optimization and explain why the transformed data is excluded during the finetuning phase.
When compared to directly using the inner-loop gradients to update the model, bilevel optimization results in a $3.4\%$ improvement in the cumulative return on the CSI dataset ($1.84$ vs. $1.78$) and a $12.1\%$ improvement on NASDAQ ($1.58$ vs. $1.41$).
Furthermore, incorporating data transformations during the finetuning phase leads to a noticeable performance drop. This is expected, as the transformed data may not align with the recent dynamic patterns close to the test set.

\begin{figure}[t]
\begin{center}
\centerline{\includegraphics[width=\columnwidth]{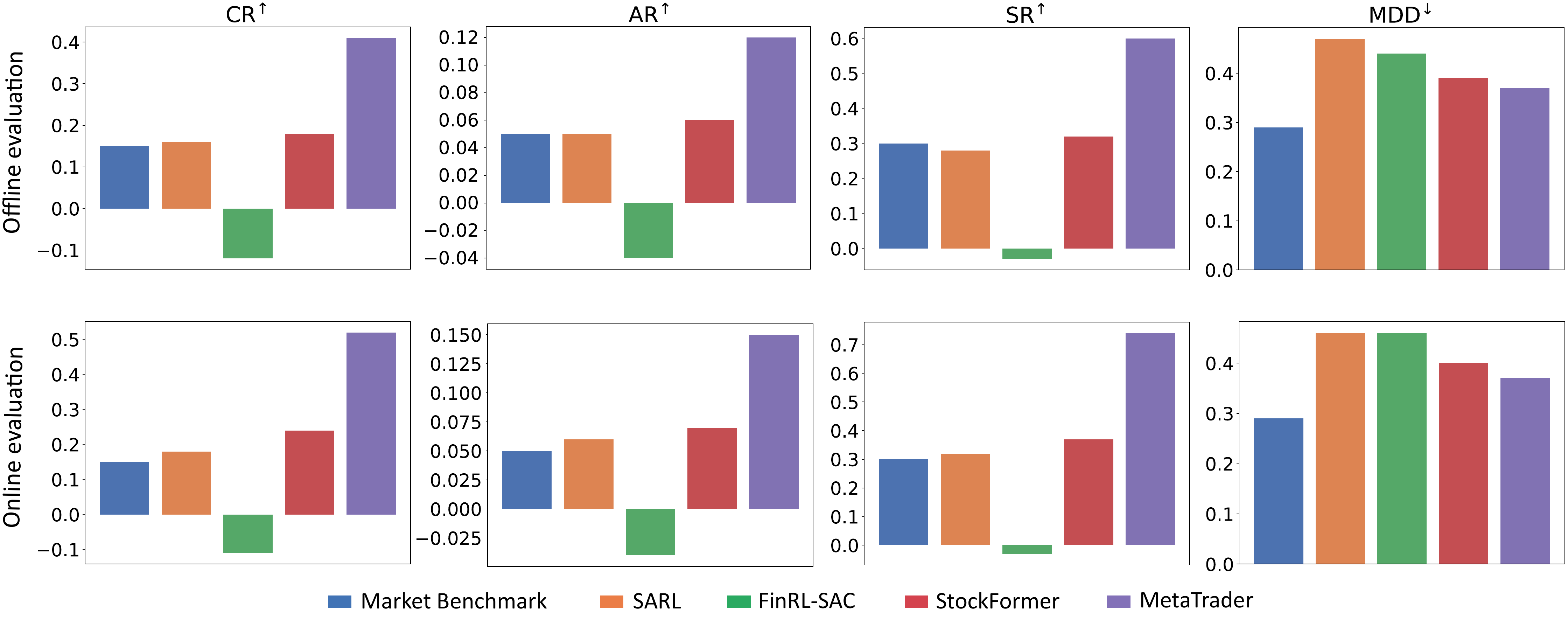}}
\vspace{-10pt}
\caption{\textbf{Experiments on the expanded dataset with $587$ stocks.}
We follow the same offline evaluation and online adaptation setups and demonstrate that \modelname{} achieves more significant performance gains over existing RL-based trading methods than those on small sets.
}
\end{center}
\label{fig:largeSet}
\vspace{-20pt}
\end{figure}

\vspace{-5pt}
\paragraph{Handling the challenges for larger market data.}

We conduct experiments on a larger dataset by expanding the range of CSI stocks and selecting a dataset containing $587$ stocks. 
%
%
Existing RL-based stock trading methods, such as FinRL, StockFormer, and SARL, primarily conduct experiments on relatively small-scale datasets. We attribute this limitation to two main factors.
From a data perspective, trading suspensions frequently occur in real-world stock data. Previous studies often select stocks based on the requirement that the proportion of valid data exceeds a specific threshold (\textit{e.g.}, $98\%$ in StockFormer) to reduce noise from excessive data interpolation.
From an algorithm perspective, as the stock pool size increases, the action space grows significantly, making it more challenging for RL methods to manage. If we aim to trade thousands of stocks in the market, the dimensionality of the action space can be even larger than the number of training sequences. The difficulty of high-dimensional action space is well-documented in other domains beyond stock trading~\cite{tavakoli2018action,saito2024potec}.

\vspace{-5pt}
\paragraph{Computational costs.} In Table \ref{tab:computational_cost}, we present the total training time and the per-sequence inference time of the compared models using a single NVIDIA RTX 3090 GPU. 
Given that our work primarily focuses on daily-level stock trading, the increased training cost introduced by bilevel optimization is acceptable, while the inference time adequately meets the efficiency demands in this scenario.

\begin{table}[t]
  \centering
  \caption{\textbf{A comparison of computational costs.} \modelname{} achieves comparable runtime efficiency to StockFormer on the CSI dataset, but with a slight increase in training time due to the bilevel learning scheme. Both models use the same pretrained feature extractors for the market states.}
   \vspace{5pt}
  \setlength\tabcolsep{20pt}
  \small
    \begin{tabular}{lcc}
    \toprule
    Method & Training time & Inference time per sequence \\
    \midrule
    StockFormer  & 28min 02s & 19.03ms  \\
    MetaTrader & 37min 27s & 19.06ms  \\
    \bottomrule
    \end{tabular}
  \label{tab:computational_cost}
\end{table}

\vspace{-5pt}
\paragraph{Additional model analyses.}
We provide additional experimental results in Supplementary Section \textcolor{blue}{S5},
including: (1) Comparisons of finetuning RL-based trading models, showing that the proposed bilevel RL approach improves domain adaptation by enabling effective model finetuning; 
(2) Evaluations of these approaches on more recent data, demonstrating the effectiveness of \modelname{} on test data with more pronounced distributional shifts from the training set;
(3) Evaluations of baseline models trained with an increased number of gradient steps, showing that the benefits of the bilevel RL approach do not stem from a simple increase in optimization steps.
\section{Related Work}

\paragraph{Deep learning-based portfolio optimization methods.}

There are two primary categories of deep learning-based approaches for portfolio optimization.
The first category leverages the temporal modeling capabilities of existing models to forecast asset prices~\cite{li2018stock, xu2018stock,feng2019temporal, wang2021hierarchical,duan2022factorvae, zheng2023deep}.
For stock trading, these methods are typically combined with relatively simple trading policies, such as buying stocks predicted to yield the highest returns and selling them at a predetermined time.
The second line of work employs deep RL, framing portfolio optimization as a Markov Decision Process (MDP) to make dynamic decisions about the timing and quantity of investments~\cite{deng2016deep,briola2021deep,jeong2019improving,liu2021finrl,kumar2023deep,liu2022finrl,gao2023stockformer}.
In this paper, we reassess the fidelity of these approaches, demonstrating that policies, constrained by offline state exploration, tend to memorize only the optimal strategy derived from the offline data. This limits the agent's ability to generalize to OOD data scenarios across varying market conditions.

\vspace{-5pt}
\paragraph{Bilevel optimization.}

Bilevel optimization-based meta-learning has emerged as a powerful tool for addressing various machine learning problems, such as few-shot learning~\cite{antoniou2018train,li2019revisiting,triantafillou2019meta,day2022attentional,cheng2023meta} and domain adaptation~\cite{schmidhuber1987evolutionary,DBLP:phd/basesearch/Finn18,hospedales2021meta}.
In the context of RL, it has been applied to learning dynamics models~\cite{saemundsson2018meta,nagabandi2018learning} or directly optimizing policies~\cite{duan2016rl,mishra2017simple,finn2017model,nagabandi2018learning,gupta2018meta,humplik2019meta,mitchell2021offline,pong2022offline,tang2022biased,greenberg2023train,gao2023context,ma2023metabox,wang2023offline}.
These optimization-based meta-learning models have demonstrated the potential to improve the generalizability of RL policies.
In contrast to prior work, we focus on the challenges posed by limited and non-stationary financial data in policy learning. To address these challenges, we introduce a novel bilevel RL approach that enhances policy generalizability while mitigating the issue of value overestimation.

\vspace{-5pt}
\paragraph{Ensemble RL.}

Standard ensemble RL is a technique where multiple models or agents are trained together to improve performance and robustness in RL tasks. 
The idea is to leverage the diversity of multiple learned policies or value functions to make better decisions, reduce overfitting, and enhance exploration. 
Ensemble methods are particularly useful in environments with high uncertainty, or when the goal is to learn an optimal policy from a static dataset previously collected, particularly when applied to OOD data points.
For instance, a prominent class of ensemble-based methods relies on \textit{model diversity}, where multiple Q-networks are trained to approximate the Bellman target by taking the minimum or mean value across several parallel Q-networks~\cite{an2021uncertainty,lee2022offline,wu2021aggressive,zhao2023ensemble}. 
In contrast, our method leverages \textit{data diversity} by performing ensemble bootstrapping on the Q-function using various data transformations. This approach captures a wider range of variability in the decision-making process while preserving model efficiency.

\section{Conclusions, Limitations, and Broader Impacts}

This paper presents \modelname{}, an RL method that formulates sequential portfolio optimization as a partial-offline RL problem with decoupled market states and balance states.
\modelname{} improves the model's generalizability to non-stationary stock data by integrating carefully designed stock augmentation techniques in a bilevel policy learning framework.
Additionally, we proposed a novel Q-learning method with a data transformation-based TD bootstrapping method, which aims to produce more conservative policies in highly dynamic data scenarios with limited training data points.
Experiments on two public stock datasets demonstrate the effectiveness of \modelname{} compared to existing RL-for-finance approaches, showcasing its great potential in dealing with rapidly changing financial markets.


An unresolved problem in this study is the training instability. 
Compared to stock prediction methods like FactorVAE and HATR, we observed that RL-based methods (including SARL, StockFormer, and our approach) generally exhibit larger standard deviations in performance across multiple training runs with random seeds. 
%
To address this, we plan to investigate alternative training strategies, including more robust initialization techniques and the integration of regularization methods that can mitigate the impact of random fluctuations and improve stability across different training runs.
Another limitation is that our approach is trained and validated solely on daily-level stock data, with an inference time of approximately $20$ milliseconds per sequence. We plan to enhance its computational efficiency and apply it to high-frequency trading scenarios in the future.

Our \modelname{} framework, although initially designed for financial decision-making tasks, offers principles that can be effectively extended to other decision-making domains, such as autonomous driving.
Similar to stock market dynamics, autonomous driving systems must continually adapt to an evolving environment (\textit{e.g.}, fluctuating traffic patterns). 
The state space an also be divided into action-free and action-dependent components: 
(1) The action-free traffic state represents the road conditions, including the behavior of other vehicles and external factors like weather and traffic observations that are unrelated to the agent's actions.
(2) The action-dependent embodied state represents the agent’s embodied states, such as the vehicle’s positions and speeds. The transition of this state branch depends on the actions taken by the vehicle.
Building on this decoupled MDP, we can formulate the learning problem driving agent from a previously collected traffic dataset as a partial-offline RL problem!
By leveraging modern ``World Models'', it is possible to generate new driving scenarios by transforming action-free traffic states and simulating changes in road conditions or the behavior of other vehicles, such as altering the trajectories of other vehicles, introducing random road disruptions, or simulating unexpected weather changes.
By implementing bilevel RL with an ensemble-based TD target on the transformed data, we can train a driving agent using the proposed algorithm, ensuring that autonomous systems can handle the complexities of dynamic, non-stationary traffic environments.

\section*{Acknowledgments}

This work was supported by the National Natural Science Foundation of China (Grant 62250062), the Smart Grid National Science and Technology Major Project (Grant 2024ZD0801200), the Shanghai Municipal Science and Technology Major Project (Grant 2021SHZDZX0102), and the Fundamental Research Funds for the Central Universities. 








\bibliographystyle{apalike}
\bibliography{sample}



\newpage

\section*{Supplementary Information}

\section*{S1 \quad Data Transformation}
\label{apdx:example}

We transform the data in sequences of $64$ days in length to construct the subsets. While the data transformation techniques are briefly illustrated in the main text, we present more details of the implementation here.

Consider a specific stock $A$: It provides an input sequence to the model, in which the daily closing prices can be denoted as $O_{F_0}^\text{close}=\{o_0^\text{close}, o_1^\text{close},\ldots,o_{63}^\text{close}\}$. 
The subsequent prices after this sequence are $o_{64}^\text{close},o_{65}^\text{close},\ldots$, and so forth.
Accordingly, we have the sequence of growth rate between daily closing prices for this stock: $\Delta O_{F_0}^\text{close}=\{0, \Delta o_1^\text{close},\ldots,\Delta o_{63}^\text{close}\}$. For example, $o_9^\text{close}=o_8^\text{close}\times (1+\Delta o_9^\text{close})$.
Without loss of generality, let us assume that its daily growth rate on the $10$-th day (\textit{i.e.}, $\Delta o_9^\text{close}$) is among the Top-$10\%$ within the stock pool.


In the first transformation method, the price sequence of stock $A$ is transformed into another sequence denoted by $O_{F_1}^\text{close}=\{o_0^\text{close},o_1^\text{close},\ldots,o_8^\text{close},o_9^\prime,o_{10}^\prime,o_{11}^\prime,\ldots,o_{63}^\prime\}$, where $o_9^\prime=o_8^\text{close} \times (1-\Delta o_9^\text{close})$. We retain the daily price growth rates on other days, such that $\Delta o_{t}^\prime = \Delta o_t^\text{price}$ for $t \geq 10$.
%
%
%
In particular, on days when the number of stocks with positive price growth does not reach $10\%$ of the total, only those stocks with positive growth will have inverted growth rates. It is noteworthy that although we manipulate the input data by measuring the daily closing prices only, we also transform the open/high/low prices along with the closing prices, while keeping the original data for the trading volumes unchanged.

In the second transformation method, the original price sequence of stock $A$ is reversed to construct another sequence of $O_{F_2}=\{o_{63}^\text{close},o_{62}^\text{close},\ldots,o_{0}^\text{close}\}$.

In the third transformation method, the price sequence is transformed into \sloppy $O_{F_3}=\{o_0^\text{close}, o_4^\text{close},o_8^\text{close},\ldots,o_{248}^\text{close},o_{252}^\text{close}\}$.

For transformations $F_2$ and $F_3$, all input data (high/low/volume) will be shifted alongside corresponding price data. For all data transformation methods, we carefully divided the training and test sets based on dates, ensuring that all transformations were applied exclusively to the training set. This guarantees no data leakage and ensures a fair comparison among all methods.

\section*{S2 \quad Datasets}
\label{apdx:datasets}

\paragraph{Data preprocessing.}
For the CSI-300 stock dataset, we follow previous work \cite{feng2019temporal,gao2023stockformer} to retain the stocks that have been traded on more
than $98\%$ training days since 01/17/2011.
For the NASDAQ-100 dataset, we also use the $98\%$ criteria to filter stocks, which derives an investment pool of $86$ stocks. 
If a stock is suspended from trading, we interpolate the missing training data using the daily changing rate of CSI-300 Composite Index or the NASDAQ-100 Index.





\paragraph{Data normalization.}
We perform normalization separately for each stock, ensuring that all normalization factors are specific to the data of the individual stock. 
%
%
For a given stock, all price data (open, close, high, low) share the same normalization factor. The normalized values can be formulated as 
\begin{equation}
N_{t_i}^{\text{price}} = \frac{o_{t_i}^{\text{price}} - \min\limits_t\{o_t^{\text{low}}\}}{\max\limits_t\{o_t^{\text{high}}\} - \min\limits_t\{o_t^{\text{low}}\}},
\end{equation}

The normalization for volume is expressed as
\begin{equation}
N_{t_i}^{\text{volume}} = \frac{o_{t_i}^{\text{volume}} - \min\limits_t\{o_t^{\text{volume}}\}}{\max\limits_t\{o_t^{\text{volume}}\} - \min\limits_t\{o_t^{\text{volume}}\}},
\end{equation}
where $N_{t_i}$ represents the normalized value, and the superscript "price" refers to the four price data types: open, close, high, and low.

\paragraph{Technical indicators.}
To align with StockFormer~\cite{gao2023stockformer}, we use the Stockstats package to compute the technical indicators listed in \cref{tab:technical_indicator}, which are incorporated as part of the observation data in our work.

\begin{table}[t]
\caption{\textbf{Technical indicators and descriptions.} Similar to StockFormer~\cite{gao2023stockformer}, these technical indicators are used as parts of the input observation data in \modelname{}. }
\label{tab:technical_indicator}
\vspace{5pt}
\centering
\renewcommand\arraystretch{1.5}
\begin{tabular}{c|c}
\toprule
Technical Indicator & Description\\ 
\midrule
macd & Moving average convergence divergence \\ 
boll\_ub & Bollinger bands (upper band) \\ 
boll\_lb & Bollinger bands (lower band) \\ 
rsi\_30 & 30 periods relative strength index \\ 
cci\_30 & Retrieves the 30 periods commodity channel index \\ 
dx\_30 & Directional index with a window length of 30 \\ 
close\_30\_sma & 30 periods simple moving average of the close price \\ 
close\_60\_sma & 60 periods simple moving average of the close price \\ 
\bottomrule
\end{tabular}
\end{table}

\section*{S3 \quad Hyperparameters}
\label{apdx:hyperparameter}

In \cref{table:hyperparameters}, we provide the hyperparameter details in both the OOD policy learning phase and the in-domain finetuning phase. For the feature extraction module, we adopt the identical hyperparameters as those employed in StockFormer~\cite{gao2023stockformer}.

\begin{table}[h]
\caption{\textbf{Training hyperparameters of \modelname{}.} These hyperparameters are used in both the OOD policy learning phase and the in-domain finetuning phase.
} 
\vspace{5pt}
\label{table:hyperparameters}
\renewcommand\arraystretch{1.5}
\centering
\begin{tabular}{c|c|c}
\toprule
Notation           & Hyperparameter     & Description           \\ 
\midrule
$\eta_1$      & 0.00001  & learning rate of the critic (inner loop) \\ 
$\eta_2$       &  0.0001  & learning rate of the critic (outer loop)    \\ 
$\alpha_1$  & 0.00001   &    learning rate of the actor (inner loop) \\ 
 $\alpha_2$     & 0.0001 & learning rate of the actor (outer loop)    \\ 
$d_\text{hidden}^1$           &  256      &number of MLP channels in the critic              \\ 
$d_\text{hidden}^2$           &  256      & number of MLP channels in the actor               \\ 
 $B, K$        & 32    & batch size, number of sampled subsets per iteration   \\ 
$M$      & 216  & number of time period slices    \\ 
$N$        & 3  & number of   stock augmentation techniques    \\
$T$            &   64     & length of time period slices        \\ \bottomrule
\end{tabular}
\end{table}

\section*{S4 \quad Evaluation Metrics}
\label{apdx:metrics}

\paragraph{Cumulative return (CR):} This is a measure of the income generated by an investment portfolio over a specific period. Specifically, it includes the entire test period.
\begin{equation}
    \begin{gathered}
o_t^{\text {close }} \in \mathbb{R}^{|S|},\quad z_t^{\prime}=z_t^{(2:|S|+1)} \in \mathbb{R}^{|S|}\\
A_t=z_t^{\prime} \cdot o_t^{\text {close }}=\sum_{i=1}^{|S|} z_t^{\prime(i)} \cdot o_t^{\text {close }(i)}, \quad CR_t=\frac{A_t}{A_0}-1
\end{gathered}
\end{equation}    
where $A_t$ represents the total asset value at time $t$ and $A_0$ denotes the initial asset value. In practice, we assume all transactions are executed at the closing price $o_t^\text{close}$.

\paragraph{Annualized return (AR):}
This is a measure of the investment growth over one year.
\begin{equation}
        AR={CR}_t^{\frac{d}{t}}-1,
\end{equation}
where $d$ represents the total number of trading days in one year.

\paragraph{Sharpe ratio (SR):} This is a metric in finance to measure the performance of an investment compared to a risk-free asset.
\begin{equation}
    SR=\frac{CR-R_f}{\sigma_p},
\end{equation}
where $R_f$ is the risk-free rate of return. $\sigma_p$ is the standard deviation of the portfolio's excess return. For our experiments, the risk-free rate used in the analysis is set to 0.

\paragraph{Maximum drawdown (MDD):} This is a measure of the maximum observed loss from a peak to a trough of a portfolio’s value before a new peak is achieved. It quantifies the largest decline during a specific period and is expressed as a percentage of the peak value. 
\begin{equation}
\text{MDD} = \max_{t \in [1, H]} \left( \frac{P_{\text{peak}, t} - P_{t}}{P_{\text{peak}, t}} \right),
\end{equation}
where $P_{\text{peak}, t}$ represents the maximum portfolio value observed up to time $t$, and $P_t$ denotes the portfolio value at time $t$. $H$ is the total number of time steps in the evaluation period.  
MDD provides insight into the portfolio’s risk by showing the potential downside during periods of significant market declines. In practical applications, MDD helps assess the stability and robustness of an investment strategy.

\section*{S5 \quad Additional Results}
\label{sec:additionalResults}

\begin{table}[t]
\centering
\caption{\textbf{Results on the CSI dataset with more recent test data.} The test set spans from 2022-05-01 to 2024-05-01, following the offline evaluation setup.
}
\vspace{5pt}
\small
\setlength\tabcolsep{20pt}
\begin{tabular}{l|rrrr}
\toprule
Method & CR$^\uparrow$ & {AR$^\uparrow$} & SR$^\uparrow$ & {MDD$^\downarrow$} \\
\midrule
Market benchmark &$-$0.08 &$-$0.04 &0.02&0.32 \\
SARL &$-$0.13&$-$0.07&$-$0.07&0.51 \\
FinRL-SAC & 0.04& 0.01&0.03&0.49 \\
StockFormer & 0.21&0.10&0.46&0.45 \\
MetaTrader &\textbf{0.32} &\textbf{0.15}&\textbf{0.76}&\textbf{0.44} \\
\bottomrule
\end{tabular}
\label{tab:recent}
\end{table}

\paragraph{Evaluation on more recent data.}
%
%
We used data up to 2022 to ensure a fair comparison with StockFormer~\cite{gao2023stockformer}, which follows the same training and testing period division.
Moreover, we conduct additional experiments using data beyond 2022. In this experiment, we do not extend the training set range but directly test on the CSI dataset spanning from 2022-05-01 to 2024-05-01. 
As shown in \cref{tab:recent}, during this period, the overall market is weaker than that in the original test set before 2022. Consequently, the annualized returns of all methods are reduced. Nonetheless, our method consistently outperforms all baselines, highlighting its potential for profitability even under more challenging market conditions.
%

\paragraph{Effectiveness of finetuning of RL-based trading models.}
In practical RL-for-finance tasks, the naive fine-tuning approach often fails to enhance model performance on test data. This is primarily due to overfitting to specific data patterns when finetuning on more recent data. This is precisely why we propose the bilevel optimization approach for the RL method.
Theoretically, the bilevel optimization scheme can significantly enhance the model’s generalizability to new data. Similar approaches, known as model-agnostic meta-learning (MAML) ~\cite{finn2017model}, have been widely adopted to improve finetuning results in few-shot learning scenarios. Intuitively, it aims to find well-performed parameter initialization that can be quickly adapted to a new related task using only a few data and a few gradient steps.
We compare the performance of different RL methods with and without \textit{finetuning on last-year recent data}, using the same configuration as offline evaluation. We present the CR, PR, SR, and MDD on the CSI dataset in \cref{tab:finetune}. 
The results are averaged over $3$ random training seeds. Notably in the cumulative return metric, our bilevel optimization approach significantly improves the finetuning results (by $+13.39\%$), while the previous RL approaches do not support such effective model finetuning (\textit{e.g.}, by $+0.81\%$ for StockFormer).

\begin{table*}[t]
	\centering
         \caption{
        \textbf{Analyses of finetuning the models on last-year training data.} The results are averaged over $3$ random training seeds. Compared with previous methods, our bilevel RL approach facilitates effective model finetuning, with a $13.39\%$ performance gain (vs. $0.81\%$ on StockFormer). 
        }
	\vspace{5pt}
 \small
	\setlength\tabcolsep{8pt}
	\begin{tabular}{l|ccc|ccc}
		\toprule
		\multirow{2}{*}{Method} & \multicolumn{3}{c|}{w/o Finetuning} & \multicolumn{3}{c}{w/ Finetuning} \\
	     & CR$^\uparrow$ & {AR$^\uparrow$} & SR$^\uparrow$ & CR$^\uparrow$ & {AR$^\uparrow$} & SR$^\uparrow$ \\
      \midrule
	     SARL
      & 1.03$\pm$0.13&{0.27$\pm$0.03} &0.89$\pm$0.08
      & 1.06$\pm$0.14 & 0.27$\pm$0.03 & 0.98$\pm$0.08 
      \\
      CQL
      &0.69$\pm$0.05 &0.19$\pm$0.01 &0.83$\pm$0.05
      &0.64$\pm$0.07 & 0.18$\pm$0.02 & 0.75$\pm$0.05 
      \\
      IQL
      &0.96$\pm$0.10&0.25$\pm$0.02 &0.89$\pm$0.04
      &1.02$\pm$0.10 & 0.26$\pm$0.02 & 0.94$\pm$0.06
      \\
      FinRL-SAC
      &0.80$\pm$0.07&0.22$\pm$0.02 &0.82$\pm$0.05
      &0.83$\pm$0.05 & 0.22$\pm$0.01 & 0.92$\pm$0.04
      \\
      FinRL-DDPG
      &0.63$\pm$0.13 &0.18$\pm$0.04&0.77$\pm$0.09
      &0.58$\pm$0.15 &0.16$\pm$0.04  & 0.73$\pm$0.12
      \\
      StockFormer
      & 1.23$\pm$0.09& 0.31$\pm$0.02& {1.18$\pm$0.05} 
      & {1.24$\pm$0.10} &  {0.31$\pm$0.02}& {1.20$\pm$0.06}
      \\
      \modelname{} 
      & 1.27$\pm$0.08 & 0.31$\pm$0.02 & 1.21$\pm$0.05
      & \textbf{1.44$\pm$0.07}& \textbf{0.35$\pm$0.02} & \textbf{1.35$\pm$0.08} 
      \\
		\bottomrule
	\end{tabular}
	\label{tab:finetune}
\end{table*}

\paragraph{Additional gradient steps for the baselines.} 

As our model is optimized for $30k$ steps during OOD policy learning and for $5k$ steps during model finetuning, we increase the training steps of other compared models to $35k \times 2$ and $35k \times 2 \times K$ steps respectively, where $K$ corresponds to the number of sampled subsets in each bilevel optimization step in our method. 
We can see from Table \ref{tab:steps} that after convergence, continuing training does not yield significant improvements for the baseline models.

\begin{table}[h]
\centering
\caption{\textbf{Experiments with a larger number of optimization steps.} The results are obtained on the CSI dataset under the offline evaluation setup over $3$ random seeds.}
\vspace{5pt}
\small
\setlength\tabcolsep{8pt}
\begin{tabular}{l|lccc|lccc}
\toprule
Method & Optim. steps & {CR}$^\uparrow$ & {AR$^\uparrow$} & SR$^\uparrow$ & Optim. steps & {CR}$^\uparrow$ & {AR$^\uparrow$} & SR$^\uparrow$\\
\midrule
SARL & 
$35k \times 2$ &1.01 & 0.26 &0.95 & 
$35k \times 64$ & 1.04&0.27&0.99 \\
FinRL-SAC & 
$35k \times 2$ &0.86 & 0.23 &0.94 & 
$35k \times 64$ & 0.89&0.24&0.93 \\
FinRL-DDPG & 
$35k \times 2$ &0.63 & 0.18 &0.77 & 
$35k \times 64$ & 0.65&0.18&0.79 \\
StockFormer & 
$35k \times 2$ &1.26 & 0.31 &1.21 & 
$35k \times 64$ & 1.28&0.32&1.24 \\
MetaTrader & 
$35k$ &\textbf{1.44}  & \textbf{0.35} & \textbf{1.35} & 
- & - & - & - \\
\bottomrule
\end{tabular}
\label{tab:steps}
\end{table}


\end{document}